\documentclass[11pt,a4paper,final]{article}
\usepackage[left=1in,right=1in,top=1in,bottom=1in]{geometry}
\usepackage{palatino} 
\usepackage{authblk}
\PassOptionsToPackage{sort&compress}{natbib}
\usepackage[]{natbib}

\usepackage{booktabs} % For formal tables
\usepackage[ruled]{algorithm2e} % For algorithms
\usepackage{tikz}
\usetikzlibrary{arrows}
\usetikzlibrary{shapes.multipart}
\usepackage{longtable}
\usepackage{caption}

\SetAlFnt{\small}
\SetAlCapFnt{\small}
\SetAlCapNameFnt{\small}
\SetAlCapHSkip{0pt}
\IncMargin{-\parindent}
\usepackage[all]{nowidow}

\usepackage[font=small,labelfont=bf]{caption}
\usepackage[utf8]{inputenc} % allow utf-8 input
\usepackage[T1]{fontenc}    % use 8-bit T1 fonts
\usepackage{hyperref}       % hyperlinks
\hypersetup{
    colorlinks=true,
    linkcolor=purple,
    filecolor=purple,
    citecolor=purple,      
    urlcolor=purple,
    pdfpagemode=FullScreen,
    }
\usepackage{url}            % simple URL typesetting
\usepackage{booktabs}       % professional-quality tables
\usepackage{amsfonts}       % blackboard math symbols
\usepackage{nicefrac}       % compact symbols for 1/2, etc.
\usepackage{microtype}      % microtypography

\usepackage{amsthm}
\usepackage{caption}
\usepackage{subcaption}
\usepackage{graphicx}

\usepackage{tocloft}
\usepackage{titletoc}
\usepackage{inconsolata}
\usepackage{microtype}
\usepackage{csquotes}
\SetTracking[spacing={50*,50*}]{encoding=T1, family=ttdefault}{50} % gentle squeeze
\usepackage{mdframed}
\usepackage{amsmath}

\usepackage{xcolor}
 
\definecolor{toolInlineLLM}{RGB}{31, 58, 117}      % dark navy blue: GitHub Copilot, Lean Copilot, lean-aide
\definecolor{toolChatbot}{RGB}{255, 149, 0}        % gold/amber: ChatGPT, Claude, Gemini, GitHub Copilot Chat
\definecolor{toolSemanticSearch}{RGB}{46, 125, 50}  % green: LeanSearch, Moogle
\definecolor{toolLLMOther}{RGB}{156, 110, 188}      % purple/lilac: Kimina autoformalizer, Kimina Prover
 
\newcommand{\toolInline}[1]{\textcolor{toolInlineLLM}{\textbf{#1}}}
\newcommand{\toolChat}[1]{\textcolor{toolChatbot}{\textbf{#1}}}
\newcommand{\toolSearch}[1]{\textcolor{toolSemanticSearch}{\textbf{#1}}}
\newcommand{\toolOther}[1]{\textcolor{toolLLMOther}{\textbf{#1}}}
\usepackage{mathtools}
\usepackage{amsthm}
\usepackage{amsmath}
\usepackage{amssymb}
\usepackage{bbm}
\usepackage{tikz} 
\usepackage{wrapfig}
\usepackage{floatrow}
\usepackage{enumitem}

\usepackage[suppress]{color-edits}
\usepackage{caption}
\usepackage{subcaption}

\usepackage{xcolor}

\newcommand{\cofirst}{\textsuperscript{\rmfamily*}}
\newcommand{\cosenior}{\textsuperscript{\rmfamily\textdagger}}

\title{Human agency in initial human-AI proof formalization workflows}

\begin{document}

\author[a,b,c]{Katherine M. Collins\cofirst}
\author[d]{Simon Frieder\cofirst}
\author[b]{Jonas Bayer}
\author[b]{Jacob Loader}
\author[e]{Jeck Lim}
\author[e,f]{Peiyang Song}
\author[a]{Fabian Zaiser}
\author[c]{Lexin Zhou}
\author[f]{Shanda Li}
\author[d]{Sam Looi}
\author[a]{Joshua B. Tenenbaum}
\author[b]{Umang Bhatt}
\author[b]{Adrian Weller}
\author[g,b]{Jose Hernandez-Orallo}
\author[a]{Cameron E. Freer\cosenior}
\author[f]{Valerie Chen\cosenior}
\author[h]{Ilia Sucholutsky\cosenior}

\affil[a]{Massachusetts Institute of Technology}
\affil[b]{University of Cambridge}
\affil[c]{Princeton University}
\affil[d]{University of Oxford}
\affil[e]{Caltech}
\affil[f]{Carnegie Mellon University}
\affil[g]{Universitat Politècnica de València}
\affil[h]{New York University}
\affil[ ]{\textsuperscript{*}Co-first author.\quad\textsuperscript{\textdagger}Co-senior author.}

\maketitle

\begin{abstract}

For centuries, human mathematicians have written proofs to substantiate their mathematical arguments; yet, the ability to automatically verify the validity of proofs has long been a challenge. Advances in AI systems' ability to generate code and engage in increasingly high-level mathematical reasoning promise to transform people's ability to formalize and thereby verify proofs. While many works focus on benchmarking the current frontier, we instead study how people use these tools and apply agency in doing so. We conduct a mixed-methods analysis into the initial impact of AI on people's formalization workflows: what people claim they want, what they see as the barriers to those visions, and how they actually use and adapt AI in practice. A qualitative survey reveals that people's preferences are diverse, but with a general desire for AI assistance in formalization that preserves high-level human control and agency over the proof discovery process. To assess how people actually engage with AI for formalization, we conduct a controlled user study in which participants formalize informal math problems and their proofs, with and without AI, across a range of mathematical problems at varying levels of difficulty and domains. Despite limitations of the tools at the time for autoformalization, participants tended to attain higher formalization accuracy when allowed access to AI tools than when formalizing on their own, with most participants flexibly choosing to use multiple different AI tools. Taken together, our work sheds light on the early stages of AI integration into formalization workflows, involving an intimate interplay of human agency and AI engagement.

\end{abstract}

Mathematics has long been a bastion of intelligence. Artificial intelligence (AI) is making tremendous strides in mathematical reasoning, attaining a gold medal in the International Math Olympiad \citep{deepmind2025imo} and even helping mathematicians crack long-standing research problems~\citep{davies2021advancing, romera2024mathematical, ellenberg2025generative, liu2025ai, epochai2026frontiermath, hariharan2026milestoneformalizationspherepacking, openai2026unitdistance}. But with all of this progress, it remains an open question of how AI may impact -- and already is impacting -- everyday mathematical practice, both among professionals and the next generation of mathematicians early in learning the trade. How and where AI will impact mathematical practice is not a function solely of the capabilities of AI systems, but where people actually \textit{want} AI to influence their workflows and how they construct and instantiate new workflows in practice.

Here, we focus on one particular part of mathematical workflows: formal theorem proving. For centuries, mathematicians have engaged in the practice of discovering and writing proofs as a way to rigorously argue for or against posited conjectures. While the practice of theorem proving has been cultivated as a way to get at and affirm ``truth,'' it has been historically hard to formally verify that proofs are correct~\citep{avigad2010understanding, hales2017formal}. This has led some to speculate that many proofs may have small errors (and even some, consequential errors~\citep{lamport2023errors}). Languages for formal mathematics (e.g., Lean~\citep{moura2021lean}, Isabelle~\citep{paulson1994isabelle}, Rocq~\citep{hugo_herbelin_2026_19256047}) have begun to streamline formal verification, helping not only scale proof verification but even enabling more scalable proof writing across large teams \citep{CommelinTopaz2024}. Yet, writing proofs in such formal languages is non-trivial and often non-intuitive~\citep{bayer2024proof}. AI for formalization promises to massively scale up formalization of human-written proofs~\citep{yang2026formal}. 

Rather than focus on benchmarking improvements in AI capabilities, we focus squarely on the interplay between people and AI systems. Specifically, we take initial steps to characterize in a more naturalistic setting people's preferences for AI integration into theorem-proving workflows and how they actually construct such workflows and allocate agency (specifically control in choice and direction) in practice. We do so via a mixed methods analysis, coupling a qualitative survey with a more controlled user study. We conducted a survey of mathematics students and researchers' preferences and self-reported usage of their AI integration: where they want AI integrated, what they feel are the blockers to those visions presently, and where they do \textit{not} want AI to be integrated. The survey was run in late 2025 giving us a glimpse -- before the surge in agentic tooling -- of where people's preferences lay on a spectrum of being willing to delegate to versus control AI use in their workflows and how they view the state (at the time of the study) of AI tools to help fulfill such visions. To understand how people actually enact such workflows in practice, we designed and ran a controlled user study, in which participants were given math problems (at various degrees of difficulty, from simple number theory problems, to problems from topology) and their proofs expressed in informal mathematics and tasked with formalizing the proofs in Lean, either with or without access to AI-based tools. We observed that participants generally attained higher formalization accuracy despite limitations of the tools at the time, achieved in part by many participants flexibly designing multi-tool workflows. Our work underscores the importance of studying not just capabilities but also people's preferences for workflows and where they would like to retain agency, what they think barriers to those workflows are, and how they actually use tools in practice relative to those perceived and actual barriers.

 \begin{figure*}[!t]
    \centering
    \includegraphics[width=0.8\linewidth]{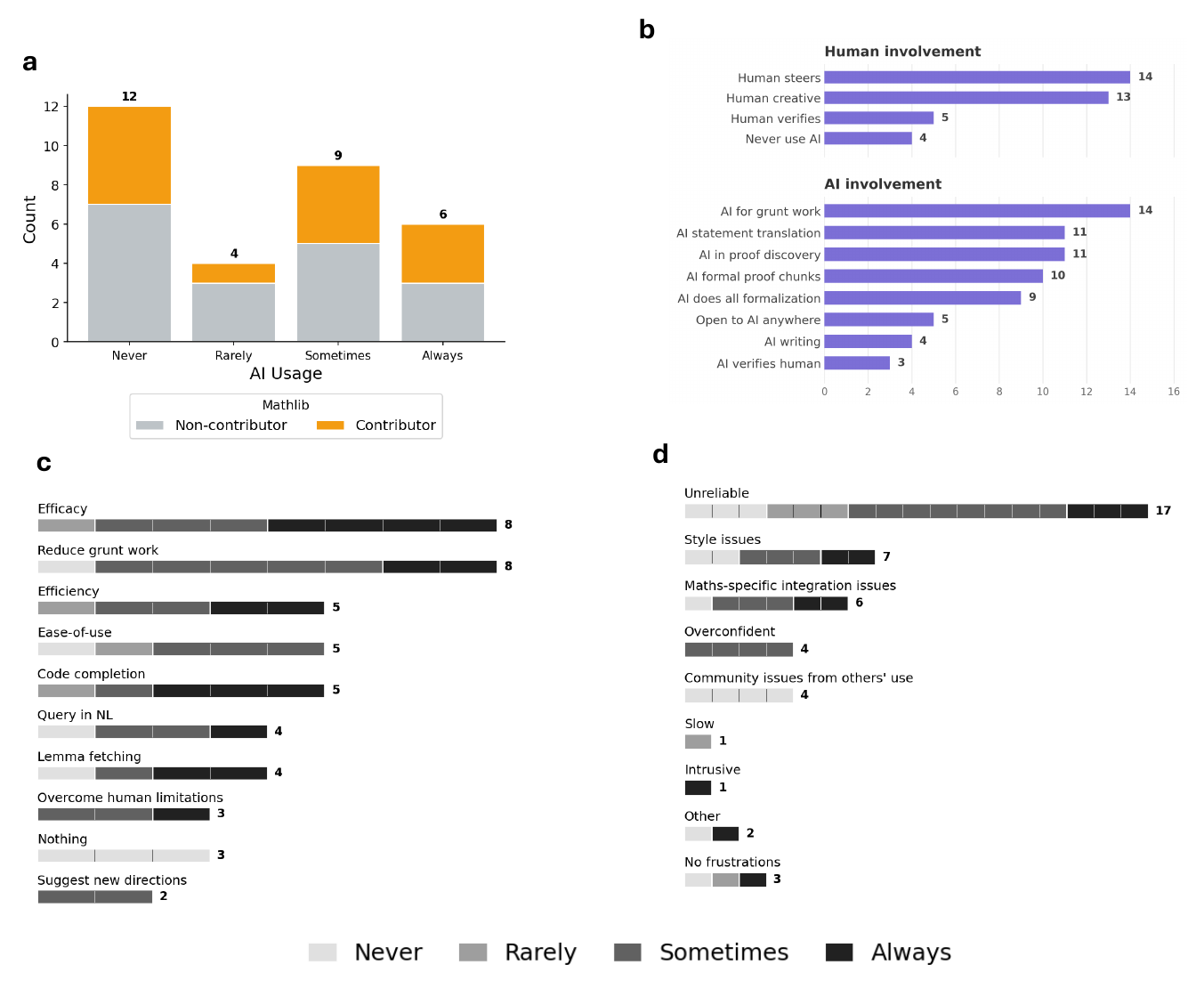}
    \caption{\textbf{Survey responses around current and envisioned AI use.} \textbf{a,} Respondents' self-reported AI usage and whether or not they contributed to the mathlib library (one indication of their experience/engagement with the Lean community); \textbf{b,} Number of survey respondents indicating preference for certain types of human and AI involvement across different aspects of formalization workflows. The human involvement coding is over the theorem proving and formalization process overall (some respondents expressed a distinction between maintaining human ``steering'' model use versus explicitly maintaining a human role in distinctively \textit{creative} tasks). AI involvement is coded more granularly over what parts of theorem proving and formalization workflows respondents specifically indicated wanting (or not wanting) AI. Note, survey responses are not mutually exclusive (the same respondent may have provided multiple codes) and survey responses were coded based on what participants said (they may fall into other categories that they did not report; see Methods). Of the $31$ respondents, $27$ responded meaningfully to this question. \textbf{c-d,} Number of survey respondents indicating a given factor in their response, for what they (\textbf{c}) liked and (\textbf{d}) were frustrated by in AI tools (at the time of the study). Responses are broken down by self-reported AI usage, and those who did not respond meaningfully to that question were excluded (of the $31$ respondents, $24$ responses were coded as meaningful in \textbf{c} compared to $27$ for \textbf{d}).} 
    \label{fig:survey-agg}
\end{figure*}

\section*{Results}

 \begin{figure*}[t!]
    \centering
    \includegraphics[width=1.0\linewidth]{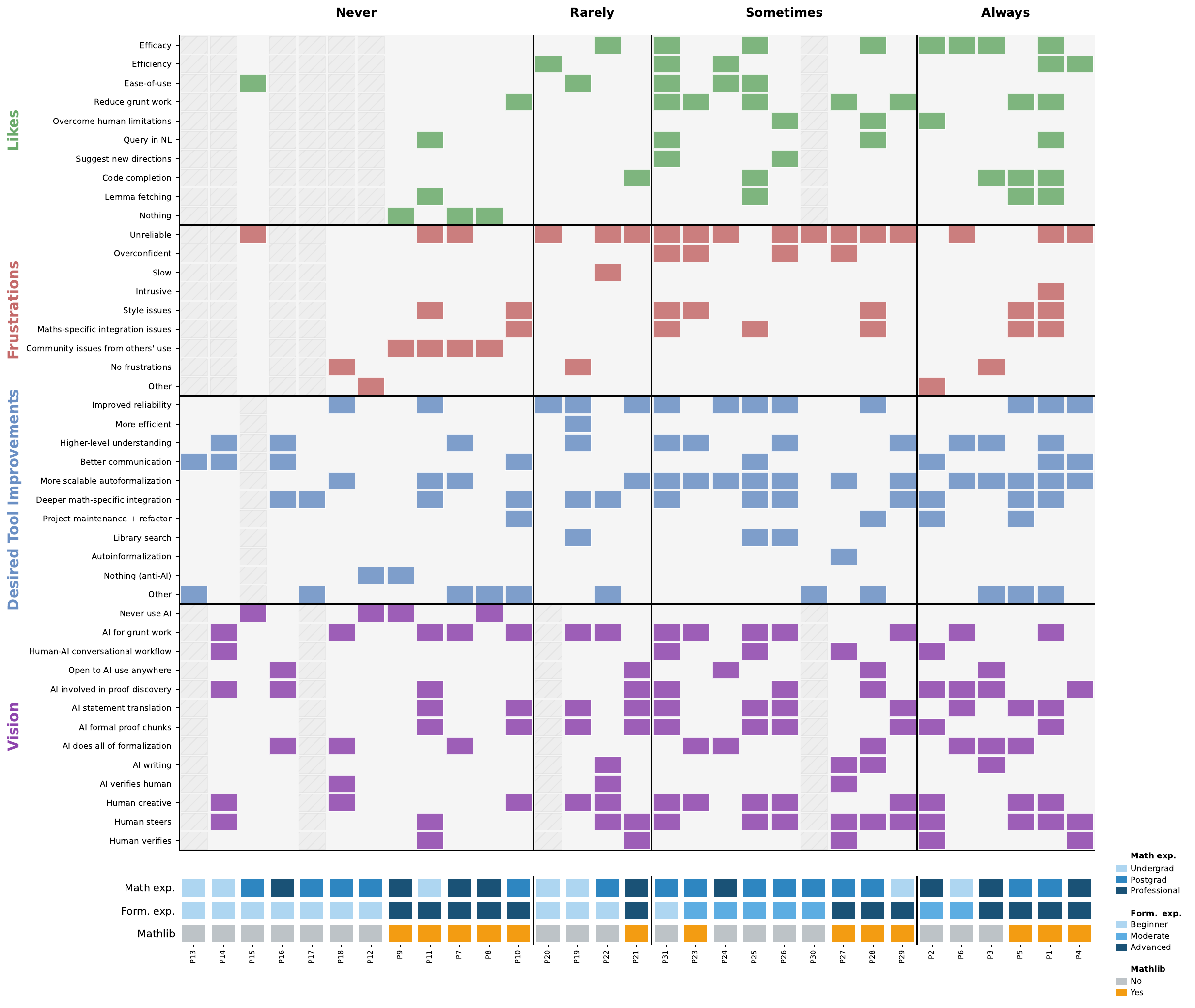}
    \caption{\textbf{Survey responses.} Full set of coded survey responses. Each column represents one respondent. A cell is colored if the respondent mentioned that factor in their response. Labels above the columns indicate the self-reported AI use of the respondent (e.g., that they ``Rarely'' used AI in their work). At the bottom, respondents' self-reported mathematics experience, formalization experience, and whether or not they have contributed to Mathlib.} 
    \label{fig:survey-agg-full}
\end{figure*}

\paragraph{People expressed heterogeneous preferences over AI use in mathematical workflows, but generally shared a desire to maintain high-level control.}

Among the $31$ survey respondents\footnote{We make all anonymized survey responses available at the following \href{https://collinskatie.github.io/hai_formalization.github.io/}{webpage}.} who passed our quality control filters, self-reported AI use varied widely and was minimally influenced by whether the respondent was an active Lean community contributor, as measured by whether they contributed to mathlib (Figure~\ref{fig:survey-agg}a) or by how much prior mathematics and formalization experience they had (see S.I.~\ref{sec:additional-survey-analyses}). 

Respondents expressed different preferences for where and how they wanted to have AI involved in mathematical practice; however, most respondents preferred workflows where AI did not fully automate mathematics (see Figure~\ref{fig:survey-agg}b). While many respondents wanted to automate core aspects of the formalization process (70.3\%), far fewer mentioned AI use in proof discovery (40.7\%) and of those who did mention AI in the formalization process, only 33.3\% indicated a desire for all aspects of the formalization process to be automated by AI. Rather, a majority of respondents (81.7\%) indicated a preference for full or at least partial human control over the process (14.8\% indicated they would never want to use AI; 66.7\% indicated a preference to maintain high-level creative and/or strategic control over the process when engaging with AI; see Figure~\ref{fig:survey-agg}b and S.I. Table~\ref{tab:envisioned-workflows}). We observe that this trend persisted across participants with different levels of AI usage and prior formalization experience (see Figure~\ref{fig:survey-agg-full}).

\paragraph{People use AI despite raising reliability concerns.}

Our survey characterized both respondents' visions for future workflows and how they actually found AI tools, at the time. While those with moderate to high use of AI generally liked that it could bring efficiency gains and reduce grunt work (Figure~\ref{fig:survey-agg}c), the dominant frustration was around lack of reliability (Figure~\ref{fig:survey-agg}d). Notably, many of the people raising concerns about reliability are those choosing to use AI (73.3\% of respondents who indicate they sometimes or always use AI expressed frustrations at the lack of reliability). Additionally, a non-negligible portion of respondents (18.5\%) mentioned formalization-specific frustrations, e.g., over the ability of AI systems they used to match their intended style of formal code as well as general frustrations of lack of integration into their specific formalization workflows (see Table~\ref{tab:envisioned-workflows}). This suggests a ripe opportunity to build more tailored AI systems for the formalization community and expand evaluation efforts beyond formal correctness.

\begin{figure*}[h!]
    \centering
    \includegraphics[width=1.0\linewidth]{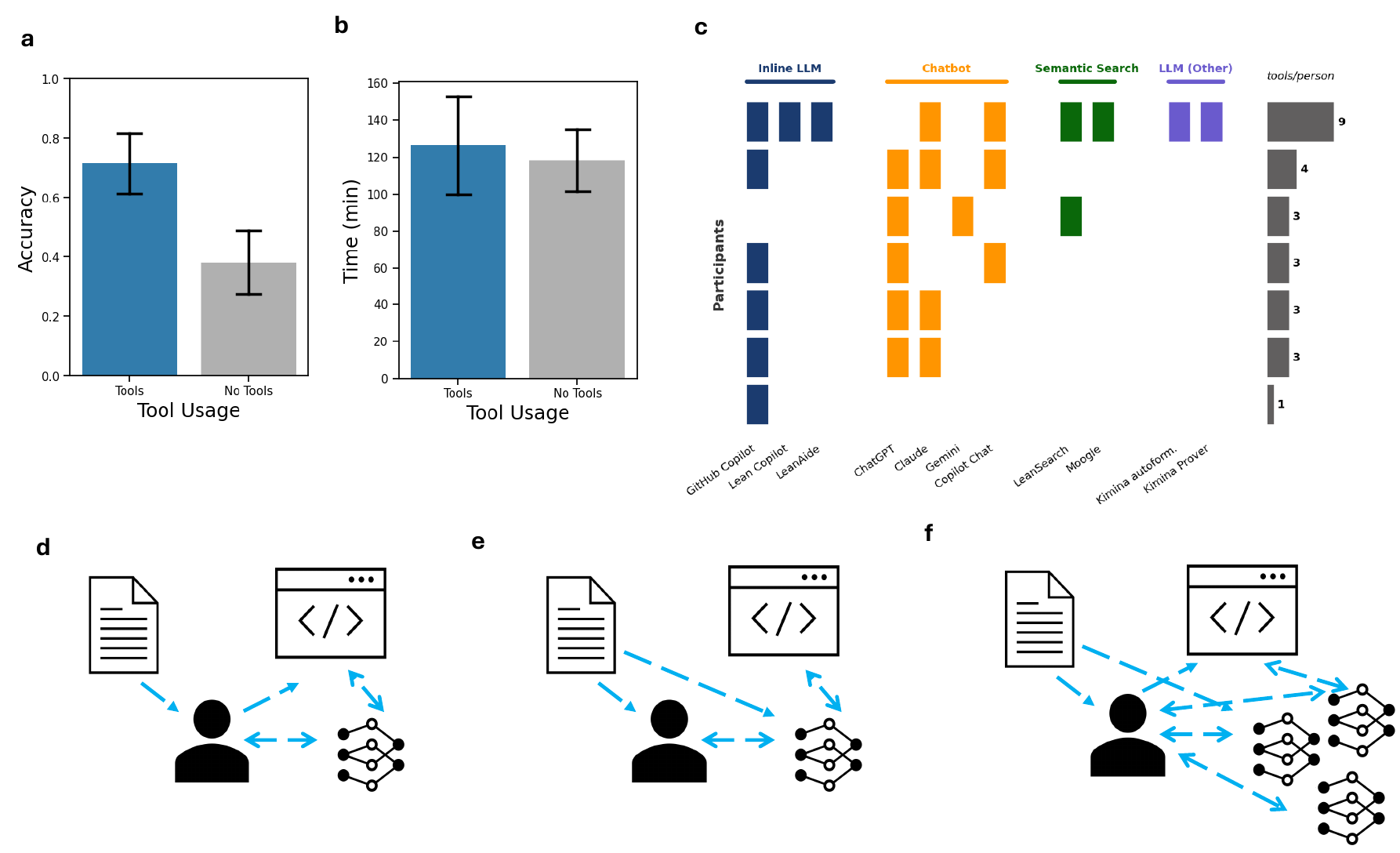}
    \caption{\textbf{Aggregate problem solving descriptive analyses, with and without tool access, and formalization workflows.} \textbf{a,} Average accuracy (both statement and proof formalized correctly, see Appendix) across problems, for groups with and without tool use; \textbf{b,} Average time taken (in minutes) across problems. Error bars depict standard error; \textbf{c,} Types of tools used by participants and estimated number of unique tools used by each participant. Each row is one participant.; \textbf{d-f} Different workflow patterns observed among participants: \textbf{d,} human formalizers with AI assistance; \textbf{e,} AI formalization with human help; \textbf{f,} varied human-AI synergistic formalization.} 
    \label{fig:acc-time}
\end{figure*}

\begin{table*}[]
\centering
 
\small
\begin{tabular}{p{16cm}}
\hline
\textbf{Self-Reported Tool Usage} \\
\hline
 
``\toolSearch{LeanSearch}/\toolSearch{Moogle}: I use these to locate relevant results in the library. My typical workflow is to use these tools via the browser, but I've gradually started using the \texttt{\#search} command from Mathlib that makes it possible to access these tools directly from the Lean editor.
\newline
\toolInline{LeanAide}/\toolOther{Kimina autoformalizer}: I use these tools to suggest ways to formalize statements in Lean. Even if the formalizations are wrong, they're still helpful because they point me to definitions in the library that I may not have previously considered.
\newline
\toolOther{Kimina Prover} model: If a lemma seems sufficiently easy, I try to offload it to the Kimina prover model to see if it can discover a proof automatically.
\newline
\toolChat{GitHub Copilot chat}: Even though most LLMs available in the GitHub Copilot chat interface are not specifically trained on Lean, I find it useful to add relevant Lean files to the chat context and ask various questions or get suggestions on planning out and structuring a proof.'' \\
\hline
 
``I used \toolChat{ChatGPT} by inputting the pdf file and asking it to formalize the proof for me. I used \toolChat{ChatGPT} by asking it to do deep research to find similar problems that have been formalized.
\newline 
I used \toolChat{Claude} by inputting the pdf file and asking it to formalize the proof for me. I used \toolChat{Claude} by asking it to do deep research to find similar problems that have been formalized. 
\newline 
I used \toolInline{Copilot} in vscode to do quick generation to small chunks of the proofs'' \\
\hline
 
``\toolSearch{leansearch.org} -- helps me find theorems that I suspect ought to already exist. This is the main thing that I was sad not to have in the NO-TOOLS segment. I describe the theorem I want in natural language or semi-formal language, and leansearch very often points me in a good direction. \toolChat{Gemini} or \toolChat{Chatgpt} -- if I'm really stuck, sometimes I'll copy the whole problem to one of these tools. Sometimes they make helpful suggestions.'' \\
\hline
 
``\toolInline{GitHub Copilot} for VSCode. I use this as intelligent autocomplete. It rarely produces correct proofs, but it can guess names and patterns very well, especially when typing out repetitive code.
\newline
\toolChat{GitHub Copilot Chat} for VSCode. I did not use during the study, but sometimes I use it in agent mode to state and prove repetitive lemmas, e.g. many trivial consequences of some theorem. I also use it for code generation, e.g. when writing Lean tactics.
\newline
Language models (e.g. \toolChat{Claude Sonnet 4}). They are helpful for finding useful Unix commands. They are not helpful for understanding mathematics because they are often confidently incorrect.'' \\
\hline
 
``I exclusively use \toolInline{GitHub Copilot} (it comes for free with my [...] account). I use Copilot to help me write normal code (e.g. C and Python) and to formalize in Lean. I usually don't use the chat feature, preferring to use the tool when I'm writing code and theorems. I tend to start function names and definitions myself, rather than writing a comment and letting the tool fill it in, which gives me greater control over the shape and feel of a project.'' \\
\hline
 
``\toolInline{Github Copilot} (all the time, for code suggestions; sometimes for questions in chat), \toolChat{ChatGPT} (sometimes, for questions)'' \\
\hline
 
``I use \toolChat{ChatGPT} to come up with general ideas about how to formalize a problem, and I use the starter code if it is any good.'' \\
\hline
 
\end{tabular}
\caption{\textbf{Participant self-reported tool usage}. After the user study, participants filled out a questionnaire. Participants were asked to indicate which, if any, AI-based tools they used in their formalization and how they used them. We include full participant responses here. Participants did not always fully report all tools they actually used in the study (see Methods and S.I.~\ref{sec:video-analyses}). Tool names are colored by category, matching Fig.~\ref{fig:acc-time}d: \toolInline{Inline LLM}, \toolChat{Chatbot}, \toolSearch{Semantic Search}, \toolOther{LLM (Other)}. Note, in some instances, e.g., the participant in the second to last row, they mention two uses of GitHub Copilot the span our colorings.} 
\label{tab:tool-use}
\end{table*}

\paragraph{Human-AI collaboration tends to increase formalization accuracy, with a mixed effect on time.} 

The survey captured respondents'  visions of what human agency might look like in future human-AI formalization workflows and people's current frustrations hindering the implementation of this vision. But what do people do in practice? Do AI tools for formalization make people more effective at formalization? And how, if at all, are they choosing to use AI to compensate for the kinds of limitations of current tools? 

\begin{figure}[h!]
    \centering
    \includegraphics[width=1.0\linewidth]{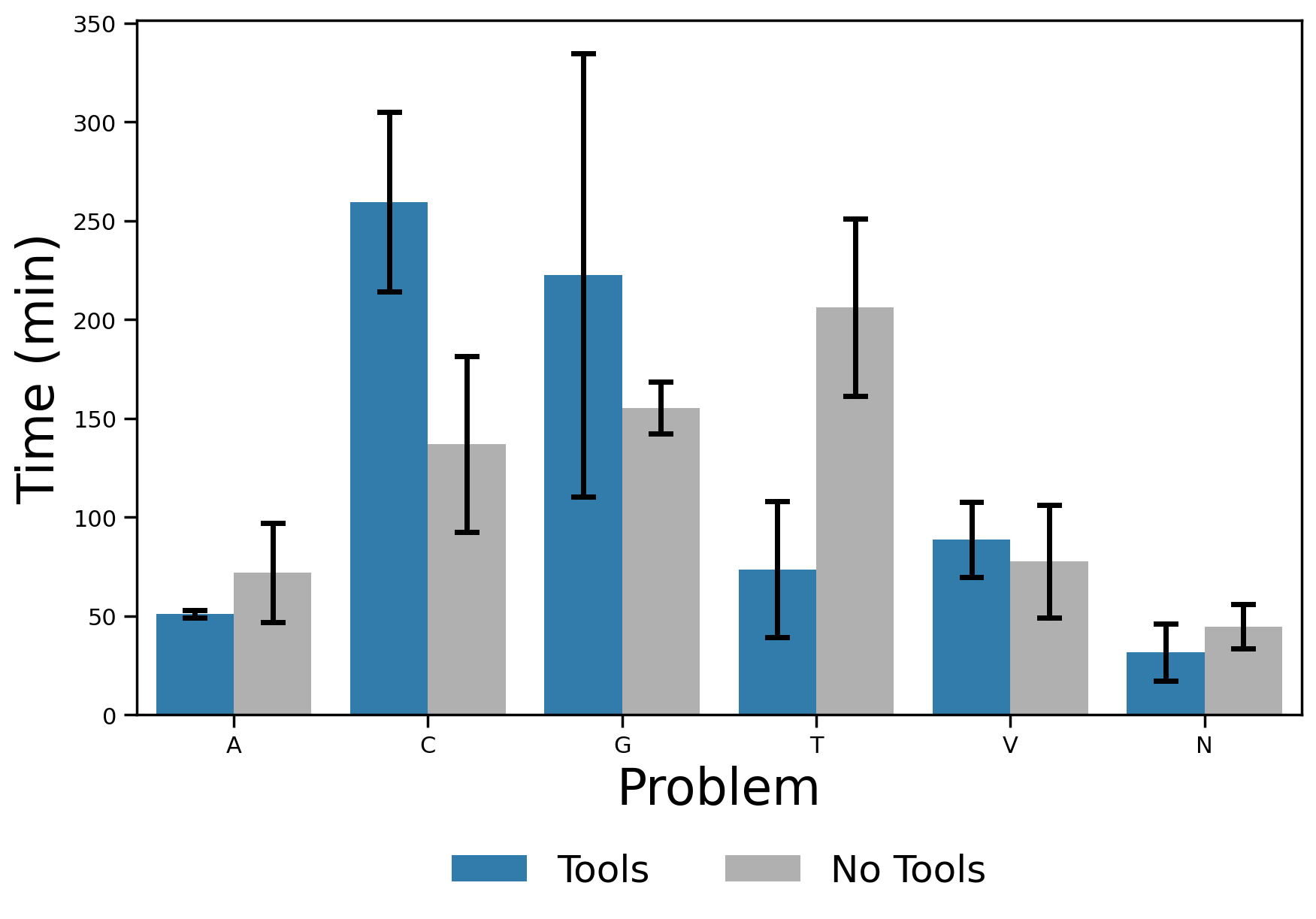}
    \caption{\textbf{Time taken per problem.} Time to formalize each problem. Error bars depict standard error.}
    \label{fig:granular-time}
\end{figure}

To take initial steps towards these questions, we ran a controlled user study. Seven participants, who ranged in formalization experience but all had at least a basic understanding of Lean and had taken at least one undergraduate level mathematics course (see Methods), were given informal proofs and tasked with formalizing both their statements and the full proof. They had six problems to formalize and were required to formalize three of the assigned problems on their own (no tools) and three with any tools they would like (see Methods). Participants' formalizations were generally more accurate when allowed tool use than when not allowed tool use (Figure~\ref{fig:acc-time}a). In a linear mixed-effects model with crossed random intercepts for participant and problem (\texttt{correct \textasciitilde{} tools + (1|participant) + (1|problem)}), tool access increased the probability of a correct response by $0.30$ points (95\% CI: $[0.10, 0.50]$; $z = 2.94$, $p = 0.003$; $n = 42$ trials from $7$ participants). Random-intercept standard deviations were $0.20$ for participant and $0.28$ for problem, indicating that problem difficulty contributed more variance than participant skill. Indeed, some problems proved highly challenging even with AI; no participant solved ``Problem G'' (see Appendix Figure~\ref{fig:acc-problem}). AI assistance did not have a statistically significant effect on time (Figure~\ref{fig:acc-time}b) and was highly dependent on the problem (Appendix Figure~\ref{fig:granular-time}). A similar linear mixed-effects model (\texttt{time \textasciitilde{} tools + (1|participant) + (1|problem)}) reveals no significant effect of tool access on time ($6.61$ 95\% CI: $[-43.62, 56.83]$; $p = 0.797$; $n = 41$ trials\footnote{One participant failed to share their video recording for one problem, hence we do not have a timestamp for that video.} from $7$ participants).

\paragraph{Most participants flexibly constructed their own multi-tool workflows.}

As in the survey, participants expressed in a post-questionnaire a range of limitations of the AI tools available at the time of the study (see S.I.~\ref{sec:participant-responses}). Participants tended to compensate for the current limitations of any one AI tool in at least two ways: using multiple AI tools and exerting prudence over AI output. Six of the seven participants opted to use more than one AI-based tool (Figure~\ref{fig:acc-time}c) and expressed deliberately tailoring their choice of tool to the task at hand (Table~\ref{tab:tool-use}).

Additionally, we conducted preliminary qualitative exploratory analysis on some of the recorded videos (see Methods) to begin to understand the actual behavior participants engage in when interacting with AI-based tools. We observed regular engagement by participants in deliberate judgment of whether to accept or reject individual suggestions (see examples in S.I. ~\ref{sec:video-analyses}). Of the videos inspected, we observed that most participants engage in the kind of behavior respondents of the survey indicated a preference for, i.e., humans maintaining strong strategic guidance over the direction of the proof. Several participants ($N=5$) engaged in most of the formalization themselves, having AI only do light autocomplete or help with lemma retrieval (we may call this group ``human formalizers with AI assistance''; Figure~\ref{fig:acc-time}d). Yet, like in the survey, we observed diversity in the actual kind of instantiated workflows. In contrast, another type (one participant) heavily used AI, only occasionally editing the proof manually (``AI formalization with human help''; Figure~\ref{fig:acc-time}e), which may more directly align with the autonomous AI desired (and at most, human verify) subset from the survey. This participant indicated limited experience in Lean and mathematics and only correctly solved one of the six problems (in the subset where AI-use was permitted). Lastly, one participant's workflow was more nuanced and intricate, involving heavy and varied AI use yet with extensive human engagement (Figure~\ref{fig:acc-time}f). This participant exemplified a more intimate ``hybrid human-AI formalization.'' They utilized many different tools across different tasks (see first row of Table~\ref{tab:tool-use}). For instance, in formalizing one problem, the participant used one tool (Lean Copilot) for suggesting lemmas, GitHub Copilot for filling in boilerplate parts of the proof, and Claude for rewriting part of the code in a different style. While this participant may have embodied the kind of pattern some respondents indicated (high-level control), the actual process required substantial switching between tools, which may not meet the desire for ease-of-use expressed by several respondents. We provide additional descriptions of example usage patterns from these initial exploratory analyses in S.I.~\ref{sec:video-analyses}.

\section*{Discussion}

Effective AI use in practice depends not only on model capabilities, but people's agency in choosing to use them. Across both the qualitative survey and user study, we saw a general preference to maintain higher-level strategic control over the process of formalization. In the user study, most participants -- even when using AI -- formalized substantial parts of the proofs themselves, relying on AI for lower-level mechanical tasks, but also continuing to engage in low-level formalization themselves. Despite the limitations of the tools of the time, many participants still found ways to attain strong performance with them. We observed that many study participants opted to use a suite of tools, tailored to particular parts of the formalization process. Likewise, participants liberally and deliberately invoked control over the suggestions provided by in-line coding tools, rejecting or deleting suggestions deemed inappropriate for the context. 

These observations underscore a core human faculty: the ability to flexibly adapt to new tools~\citep{allen2020rapid}. When assessing the impact of AI integration in human workflows -- we need to understand not only what the capabilities of tools are, but how people are and will adapt to them.  Both the survey and user study were conducted before the more recent surge of popularity in agentic coding systems \citep{li2025aiteammates}, like Claude Code \citep{anthropic2025claudecode} or Codex \citep{openai2025codex}. That we carried out the study before the rise of agentic coding systems is in some ways an advantage as it allowed us to study in detail how participants connected (manually) different tools to form workflows, which increasingly agentic coding systems could autonomously form. Our study offers a moment-in-time snapshot of initial use behavior and preferences that we cannot roll back the clock to recollect. 

Nonetheless, our study and analyses are limited by several factors. Both our survey and study involved relatively few participants and captured only a small subset of the broad space of possible mathematics students and researchers. Respondents to the survey were primarily recruited through mailing lists and the social network X, which included a diverse set of users ranging from undergraduates to professional mathematicians. A more systematic survey across levels of experience (in proof discovery and formalization) and the domain of mathematics is warranted. Our user study consisted primarily of post-graduate or postdoctoral level researchers, not expert mathematicians. We saw, at the time of the study, dramatic differences in people's willingness to use AI (in the survey and study) and the range and type of tools they engaged with if choosing to use AI — how much variability is there across a wider range of the mathematics community, and how do choices influence one another? Additionally, our instruction allowing tool use during one of the weeks may have biased participants to use AI when they otherwise may have chosen not to. 

Our user study also focuses on the formalization of given informal proofs, not the actual process of proof discovery.  It is an open and important question to more systematically understand how AI access impacts a wider gamut of mathematicians' workflows~\citep{frieder2024data} and over the kind of longer time horizons real human thought partners engage in~\citep{collins2026meaningful}. In addition to valuable new kinds of evaluations designed to probe research-level mathematics capabilities of AI systems, e.g., ~\citep{abouzaid2026proof}, we also need more structured evaluations -- a mix of qualitative case studies, of the kind in conducted in other recent works~\citep{feng2026autonomousmathematicsresearch, bubeck2025early} and more controlled (and scaled) user studies  of the kind we implement here. It is also important to move beyond just assessing accuracy. Many participants were sensitive to style and expressed desires to preserve that aspect, while automating more mechanical grunt work. This sentiment aligns with several other expert mathematicians who have shared their views on the future of mathematics, at the time of writing~\citep{tao2025machine, avigad2026mathematiciansageai}. 

Additionally, while we were able to, with consent, collect highly rich process-level data into participants' formalization and human-AI interaction behavior in the user study, careful coding of such data requires substantial analytical overhead, which we have only scratched the surface of (and are actively exploring in ongoing work). More scalable human-AI interactive video analyses, e.g., drawing on other work in human-AI interaction, are important next steps~\citep{mozannar2022reading}, not just for mathematics but the wider range of domains where we want to understand peoples' AI workflows. Here, we focus on the recordings where participants interact with AI in the tool use week, analyzing the human-only week videos can also motivate entirely new AI-assisted workflows and inform work understanding human cognition and what may make formalization challenging. 

What kinds of human-AI workflows might we expect people to engage with and create as AI tools continue to advance at a rapid rate and take on new kinds of ``agency'' of their own? In many ways, mathematics is so much more than what is proved or formalized, sitting partway between engineering, science, and artistic activities ~\citep{hadamard1954essay, hardy1992mathematician, newellCreative}. If a core part of mathematics by people for people is human understanding and the social relations that flourish around mathematics~\citep{thurston1995proof}, then AI tools which can automatically check proofs, reduce grunt work, and open up new kinds of discoveries could lay the seeds to advance human understanding. But in that quest, it is exciting -- and deeply important -- to continue to elicit and evaluate, people’s preferences and practice for human agency in the building and adaptation of new workflows, alone and potentially together with AI and \textit{other people}, for this evolving era of mathematics.

\section*{Materials  and methods}
\label{sec:methods}

\subsection*{Survey}

\paragraph{Design}

We designed a survey to better understand a wider range of people's current and preferred modes of interacting with AI as part of proof formalization and discovery. The survey was designed and conducted after the human-AI formalization user study (described below). We based our questions off of the post-survey participants in the user study answered, with modifications for clarity. We include the full questionnaire in S.I.~\ref{ref:survey-questionnaire}. The study received prior ethics approval from the University of Cambridge Department of Engineering ethics review. All respondents provided informed consent before submitting their survey.

\paragraph{Recruitment}

Participants were recruited via X, mathematics department mailing lists, and manually writing to several professors who teach classes on Lean. This survey was active between October 27, 2025, and November 25, 2025.

\paragraph{Quality Control}

Due to the wide recruitment, we received responses from several people who indicated no prior experience with, or understanding of, formalization. As our goal was to study AI use and preference around mathematical formalization, we removed any participant ($9$ of $40$ who responded) who indicated they had no experience at all with formalization or did not even know what formalization is, or did not provide substantial responses for the open-ended questions. Some respondents provided textual responses for only a few of the open-ended questions. For respondents who provided genuine responses for at least some questions, we exclude them from the analysis (listed as N/A) for questions for which they did not respond.

\paragraph{Analysis Methodology}

We focus our analyses on five primary open-ended questions, which captured what respondents liked about current AI (Q15), what they were frustrated by (Q16), what they most wanted in the next-generation of AI (Q17), and what they envisioned for future workflows (Q18-19). Full questions are included in S.I.~\ref{ref:survey-questionnaire}.  Coding of these open-ended text responses to the survey was conducted across multiple phases. First, one author read through all responses and came up with an initial coding. This coding was discussed and iterated on with three other authors (Claude Code was also consulted). One author then manually coded all responses based on these codes. Codes were then revised by these four authors with the goal of having a sufficiently rich yet relatively simple set of codes. Coding was then updated and discussed by said authors. Four other authors (not consulted in the initial coding) then coded responses. The original four authors then reconciled coding differences through discussion. The survey responses sometimes required inductive leaps of intent, hence, the extensive multi-author coding process -- which still involves subjective judgment. We make all survey responses and codings available at the following \href{https://collinskatie.github.io/hai_formalization.github.io/}{webpage}. 

We also analyze and decompose responses based on self-reported categorical features (e.g., AI use, mathematics experience) and conduct exploratory analyses into self-reported tool use (S.I. ~\ref{sec:additional-survey-analyses}).

\subsection*{User Study}

\paragraph{Problems}

We curate a selection of six problems (identified by the letters \emph{C}, \emph{N}, \emph{A}, \emph{V}, \emph{G}, \emph{T}, which loosely correspond to their mathematical domain, e.g., \emph{N} represents \emph{number theory}; see Appendix~\ref{sec:problem-design}). This relatively low number of problems is necessary to gather sufficient data for each problem and keep formalization time within reasonable bounds. Hence, we carefully source problems so that they cover a range of domains, mathematical difficulty, and formalization difficulty (since a problem being easy mathematically does not imply easy formalization), as well as contamination (to test whether participants can uncover the publicly known solution). More details on problem selection are included in Appendix~\ref{sec:problem-design}.

\paragraph{Participants}

The user study was active between May 26 and Aug 11, 2025. Eight participants originally completed our study; however, we excluded one participant who failed to record their videos separately. Participants were recruited to have at least a baseline level of Lean experience, e.g., engaged with at least a tutorial like the Natural Numbers Game~\citep{bayer2025studying}. All participants had taken at least one undergraduate mathematics level course; most had post-graduate level mathematics experience (see Appendix~\ref{sec:study-details}). In their post-survey, participants indicated their prior level of Lean experience; participants happened to be split roughly evenly along those who categorized themselves as ``Beginner'' ($N=3$) versus ``Advanced'' or ``Expert'' ($N=4$) formalizers. Participants were given six informal mathematical problems, together with their proofs, to formalize. Problems were split into two groups of three, matched by formalization difficulty across groups. Participants were given two weeks to formalize the problems and proofs; participants are instructed to formalize a group of three problems and proofs in week one, and the other group of three in week two. Participants’ tool access differed across weeks. In one week, participants were not allowed to access any tool\footnote{Participants were still allowed a text editor (e.g., VSCode) and Lean compiler.} (the ``human alone'' condition). In the other week, participants were allowed access to any AI-based online resource. This included all standard AI tools such as ChatGPT, Claude, GitHub Copilot as well as mathematics specific tools like LeanSearch for finding mathlib theorems and Lean Copilot for more tailored Lean-specific suggestions in VSCode. Participants were randomly assigned whether to use AI-assistance in week one or two. We asked that participants record their screens while working and send us the recordings. The resulting study includes over $80$ hours of participants' screen-recorded formalization processes. Additional details are in Appendix~\ref{sec:participant-rec}.

\paragraph{Analysis Methodology}
Our primary measures are formalization accuracy and time-to-formalize. Three experienced Lean users from our author team manually grade the accuracy of participants' statements and proofs; one grader marked each problem. Responses are scored with a binary correct or not based on whether both the statement and proof are correct. This is inherently subjective in our current definition of accuracy; we are actively working on expanding our accuracy evaluation for future work. We include an initial more granular analysis of correctness in Appendix~\ref{sec:acc-granular}. All our formalizations will be made publicly available upon full publication. Additionally, time is measured directly from the videos; participants were asked to record their entire problem solving session, which often unfolded over one or more sessions per problem. Time-to-formalize is summed over all sessions per problem. We also take steps to assess which AI-based formalization tools people used based on their self-report and initial exploratory analyses into a sampling of the live recorded videos. Where participants used an AI tool that they did not self-report, we did include that in the tools used as part of coding Figure~\ref{fig:acc-time}c-d. % 
%TC:ignore
\section*{Acknowledgments}

We thank Ced Zhang, Albert Jiang, Tim Gowers, Mateja Jamnik, Kaiyu Yang, Phoebe Zeng, Eric Horvitz, and Hussein Mozannar for valuable conversations that informed this work. We also thank Jeremy Avigad for kindly helping circulate our studies for recruiting participants. KMC acknowledges support from the NSF SBE SPRF, the Cambridge Trust, and King's College Cambridge. AW  acknowledges  support  from  a  Turing  AI  Fellowship  under grant  EP/V025279/1, The Alan Turing Institute, and the Leverhulme Trust via CFI. JBT acknowledges support from the AFOSR (FA9550-22-1-0387), the ONR Science of AI program (N00014-23-1-2355), a Schmidt AI2050 Fellowship, and the Siegel Family Quest for Intelligence at MIT. 

\section*{Participant Contributors}

We thank all participants of our study. The following participants opted to be listed as contributors: Xavier Lien, Cayden Codel, Mauricio Barba, and Anand Tadipatri, and Fabian Zaiser; the others elected to remain anonymous. Fabian joined our author team after submitting his study to help with analyses and the separate qualitative survey. We also thank all survey respondents. 

\section*{AI Disclaimer}

AI coding tools were used to help with data analysis and figure construction. We also used Claude to help come up with the survey coding, as described in the Methods under Analysis Methodology. We take responsibility for all results and analyses presented in this work.

\bibliography{main}
\bibliographystyle{abbrvnat}

\newpage
\setcounter{section}{0}
\renewcommand{\thesection}{A\arabic{section}}

\startcontents[appendix]
\printcontents[appendix]{l}{1}{\section*{Appendix}}

\section{Additional Related Work}

\paragraph{AI for Formalization}

Many recent formalization systems use LLMs to provide varying degrees of automation to users. Interactive tools range from retrieval-augmented generation (RAG) pipelines such as ReProver \citep{NEURIPS2023_44414694} that select relevant Mathlib premises and suggest tactics to the user, to assistants such as Lean Copilot \citep{song2024lean} and LeanProgress \citep{george2025leanprogress} that provide tactic suggestion, proof search, premise selection, and proof progress prediction natively in the editor. A greater degree of automation is provided by systems such as DeepSeek-Prover-v2 \citep{ren2025deepseekproverv2advancingformalmathematical}, Kimina Prover \citep{wang2025kimina}, or Seed-Prover \citep{chen2025seedproverdeepbroadreasoning} that are trained via reinforcement learning that aim to one-shot formalizations more accurately or make use of a range of tools to improve formalization quality. Large upstream corpora such as NuminaMath (approximately 860k competition-style problems with chain-of-thought solutions) and the Lean-specific NuminaMath-LEAN (approximately 100k human-annotated Lean 4 statements and proofs) increasingly train math LLMs and Lean provers, including Kimina-Prover, so they help contextualize the capabilities available to participants in our study \citep{li2024numinamath,numinamathlean2025,wang2025kimina}. Complementing prover systems, process-driven autoformalization in Lean 4 introduces the FormL4 benchmark and a process-supervised verifier that leverages Lean 4 compiler feedback, and Lean Workbook contributes about 57k natural language-Lean pairs via an iterative translation-and-filtering pipeline checked by the Lean compiler \citep{forml4,leanworkbook}.

\paragraph{Interactive Theorem Proving User Studies}

A body of work from the 1990s and early 2000s studied how mathematicians engage with interactive theorem proving workflows for formalization~\citep{AITKEN1998263, merriam1996evaluating, merriam1996two, kadoda2000cognitive}. For instance, in Merriam and Harrison~\citep{merriam1996evaluating} characterized four key activities that people engage in when formalizing: planning (designing the high-level sketch of the proof), reusing (leveraging previously written code and writing code strategically with reuse in mind), reflecting (evaluating what has been written and further understanding the proof goal), and articulating (actually specifying commands to the prover assistant). As we explore in our work, these four strategies can each be modularly engaged with AI-based tools for formalization (effectively or not). Additional user studies into tools for formalization have been conducted since, e.g., in de Almeida et al.~\citep{de2025lessons}, the authors analyze a wide range of Rocq (formerly Coq) users' survey responses about their use behavior, finding marked differences in styles and preferences around formalization depending on the level of experience of the user. Most closely related to our work,  Shi et al.~\citep{shi2025qed} conduct an observation study formalizing proofs in Lean and Rocq; this work offers a rich, in-depth look at modern formalization practice. The authors observe varied tool use, similar to the kind engaged in our study, and also notice that several users care about features beyond just correct formalization -- a pattern we noticed in our study as well (e.g., some participants tried to write their proofs to use particular libraries, or commented on how ``ugly'' they found their code). Most differentially from us, they do not substantively investigate peoples' interaction with AI systems as part of their formalization, in contrast to the focus of our work. More recently, several frontier labs have collaborated with mathematicians for deep dive analyses into their experiences working through one or more problems with that lab's AI systems~\citep{bubeck2025early, zheng2026ai}. We see this style of work as complementary to our more structured user study and wider ranging preference elicitation (e.g., including but not limited to earlier career mathematicians).

\paragraph{Human-AI Interaction for Programming}
A fairly extensive set of user studies have been, and continue to be, conducted to understand how developers code with various AI copilots.
These studies have focused on two forms of assistance that AI copilots provide: autocomplete 
suggestions~\citep{vaithilingam2022expectation,peng2023impact,barke2022grounded,prather2023its, mozannar2022reading, vasconcelos2023generation,cui2024productivity,mozannar2024realhumaneval} and chat dialogue~\citep{ross2023programmer, chopra2023conversational, kazemitabaar2023studying, gu2023analysts,nam2024using,mozannar2024realhumaneval}.
These studies show how copilots have generally had a positive impact on software development, e.g., leading to an increase in perceived productivity~\citep{ziegler2022productivity,10.1145/3706599.3706670} and rate of task completion in controlled studies~\citep{vaithilingam2022expectation, peng2023impact} compared to developers writing code on their own.
More recently, moving beyond copilots for software development, we have seen the introduction of coding \emph{agents} (e.g., Devin~\citep{devin}, OpenHands~\citep{wang2024openhands}, Claude Code~\citep{claudecode}).
A growing set of work measures the impact of these more autonomous tools on developer workflows~\citep{impact_software_development,chen2025code,becker2025measuring,chen2025can}. In our study, we also observe how these AI coding assistants are also increasingly integrated into mathematicians' proof formalization workflows.

\section{Additional Survey Analyses}
\label{sec:additional-survey-analyses}

\begin{table}[!t]
\centering
 
\fontsize{8.0}{10.2}\selectfont

\begin{tabular}{p{5.2cm}|p{5.2cm}|p{5.2cm}}
\hline
\textbf{Desired tool improvements} & \textbf{Desired workflows} & \textbf{Desired human vs. AI involvement} \\
\hline
 
``I would like to have a much smaller gap between what is currently written in mathematical papers and what is needed to get a proof checker to accept it as valid.'' &
``I would be happy to provide formalisation of statements and proof outlines, even to a level of detail that is (much) higher than current mathematical practice and then wait to see what the AI is able to fill in in terms of formalised proofs.'' &
``The lowest expectation would be that AI could generate all/most of the trivial results about new definitions and basic properties. Possibly, it would also suggest in real time when there is an automated way of closing the current proof. I would welcome *any* help. However, if the "help" comes at the cost of being to continuously suspicious of whether the AI managed to find and exploit a "bug", then I would probably think that I can be quicker and more productive not relying on it.'' \\
\hline
 
``It's like having a lab partner who never sleeps. When I've been staring at the same problem for an hour, it can instantly suggest a new angle I hadn't considered. It's best for those "oh right, I forgot about that tactic" moments.'' &
``A tool that learns from its own mistakes. If it suggests a tactic and Lean throws an error, it should use that error message to self-correct and offer a better suggestion, instead of me having to start a new chat.'' &
``Help me with: the tedious parts and the memory lapses. Remind me of lemma names, write the boring boilerplate. Don't help with: the core creative idea. If it just gives me a full proof I don't understand, I haven't learned anything. The "why" should still come from me.'' \\
\hline

 ``I want to hit a few keys (i.e. a keybinding or key shortcut) to have the AI make higher-level suggestions. For example, suppose I write a function for List. I'd then like to press a few keys and have the AI automatically insert lemma statements and proofs. In other words, I'd like the "agentic" AI tool to act more "agentically," supplying fully-formed thoughts (that typecheck!) when I finish a thought.'' &
``I'd love it if I can write a bunch of theorem statements and then use shorthand for the proof bodies, and then have the AI agent fill in the proofs. For example, "pf by induction using lemmas x, y, z. Branch W is probably by contradiction, and the key insight is PQR. Will likely need new lemmas B and C." And then I move on to the next thing later in the file while the agent works.'' &
``I want tools that follow my design and instructions, and not insist on "its way." I'm an expert in formalization, so I have a good sense for how to design data structures/definitions/proofs that makes it easier to maintain a project or get proofs to compile quickly. Perhaps somewhat contradictingly, I want tools to semantically handle refactors. There are many isomorphic ways to write a data structure, and as a project proceeds, it sometimes becomes clear that method B is better than method A. With formalization projects, it's harder to refactor things cleanly without breaking proofs. An AI tool that can do that for me would be great.'' \\
\hline

``Auto INformalizer! I have started to do "new" research math directly into Lean, and then to write it up is a hassle. I want an AI tool that will informalizes my Lean code in a human-readable format'' &
``Starting in Lean! Okay, drawing on a whiteboard, then straight into Lean. Once the proof is formalized, then since it is so clearly specified it *should* be easy to port it down into regular mathematical language. (Notably, this is not currently easy)'' &
``Informalization, generating code. I really worry about informalization, though. It can miss crucial steps and hallucinate others. Maybe this should be more classical symbolic methods. At least a combination of the two. Also I want a tool that can help me make a small change in my code and propagate that throughout an entire project. Like, restating a lemma to change variables to/from implicit/explicit, and updating every theorem to reflect said change'' \\
\hline
 
``I'd like future AI-based formalization tools to be far more reliable: no hallucinated lemmas, consistently idiomatic proofs, and suggestions that actually compile. I also hope they gain a deeper understanding of larger proof structures -- helping with definitions, organization, and multi-step reasoning rather than just local goals. Overall, I want a tool that feels like a dependable co-author, not something I need to constantly fix.'' &
``I imagine a workflow where I move fluidly between informal reasoning and AI-assisted formalization. I would sketch the proof idea, and the tool would propose precise definitions, outline the formal structure, and generate initial Lean code that actually compiles. From there, I'd iterate: refining the strategy, asking the tool to fill in missing steps, and letting it reorganize or simplify parts of the development. Ideally, discovering a new proof and formalizing it would become a single, integrated process rather than two separate stages.'' &
``I would want AI-based tools to help with repetitive, mechanical, or search-heavy tasks---things like looking up relevant lemmas, generating boilerplate code, suggesting small proof steps, or pointing out possible strategies when I'm stuck. These tools are most helpful when they reduce overhead without reducing my understanding or control. I would not want AI to be used for fully generating proofs of novel results or making high-level design decisions. I want to maintain ownership of the core reasoning, choices of definitions, and proof structure. Using AI in these areas risks introducing errors, misaligned approaches, or a superficial understanding of the development.'' \\
\hline
 \normalsize
\end{tabular}
\caption{\textbf{Example survey responses around envisioned formalization workflows.} Each row is one survey respondent. Responses were chosen to highlight variation across participants in some of the richer responses.}
\label{tab:envisioned-workflows}
\end{table}

\begin{table*}[]
\centering
\small
\begin{tabular}{p{16cm}}
\hline
\textbf{Self-Reported Tool Usage} \\
\hline
When I formalize, I almost always use GitHub Copilot. It helps me autocomplete blocks of boilerplate code or fill in almost-symmetric proof branches. I almost never use the chat feature of Copilot, preferring the autocomplete features that react when I type.\\
%\newline
When I write scripts and C code, I sometimes use ChatGPT (maybe once or twice a day). I use ChatGPT mainly as a smart search feature/reactive StackOverflow. Often I ask about Bash scripts and syntax or weird C/C++ behavior, such as I/O buffering or signed casting.\\
\hline
I use leansearch maybe once every few weeks, when I can't find something using loogle and the documentation and want to make sure it's really not there.\\
\hline
GPT-5 Pro for mathematical details and proof architecture\\
%\newline
Claude Code with Lean skill and Lean LSP MCP for filling sorries and fixing errors\\
\hline
Claude code is really good at semantics of programming languages. It is extremely good at converting between lambda calculi. For proofs i tell it to use grind and it gets the rest of the structure right.\\
%\newline
LeanSearch is very powerful to search mathlib for harder theorems that i cant loogle\\
\hline
Copilot: generally have it on in VSCode. It's great for making lemmas similar to ones I already have, and writing out lots of similar cases. Essentially, it helps me type less. I have not found it useful for generating totally new code with new ideas/patterns\\
%\newline 
ChatGPT: I have repeatedly tried to get it to generate Lean code, with no success. It is, however, great for helping with terminal commands and using lake. For example, I used it a lot to generate an import graph for my project. It also helped quite a bit generating code for my widget (I am not using qq, and instead using ChatGPT)\\
%\newline 
Moogle: I wish it would work better. Generally I have better success typing out the lemma I want and hitting 'apply?'\\
\hline
Only GitHub Copilot. It helps with auto-completion and in suggesting the next tactic.\\
\hline
GitHub Copilot. My use is very superficial: when it suggests an autocompletion, I sometimes accept it.  I am quite likely to accept autocompletions for documentation, but sometimes regret accepting code suggestions.\\
%\newline
ChatGPT. I have asked ChatGPT mathematical questions, but I did not trust the answer it gave me.  So, I ended up formalising what it said (and it is sometimes correct!). I never asked ChatGPT to write Lean code for me, but I can tell when my students do, since their imports are often files that do not exist and sometimes the syntax is the one of Lean 3.\\
\hline
\end{tabular}
\caption{\textbf{Example survey respondents' self-reported tool usage}. Survey respondents were asked whether they used any of the AI tools that participants reported in the user study. We include several example responses.}  
\label{tab:tool-use-survey-examples}
\end{table*}

\begin{figure}
    \centering
    \includegraphics[width=1.0\linewidth]{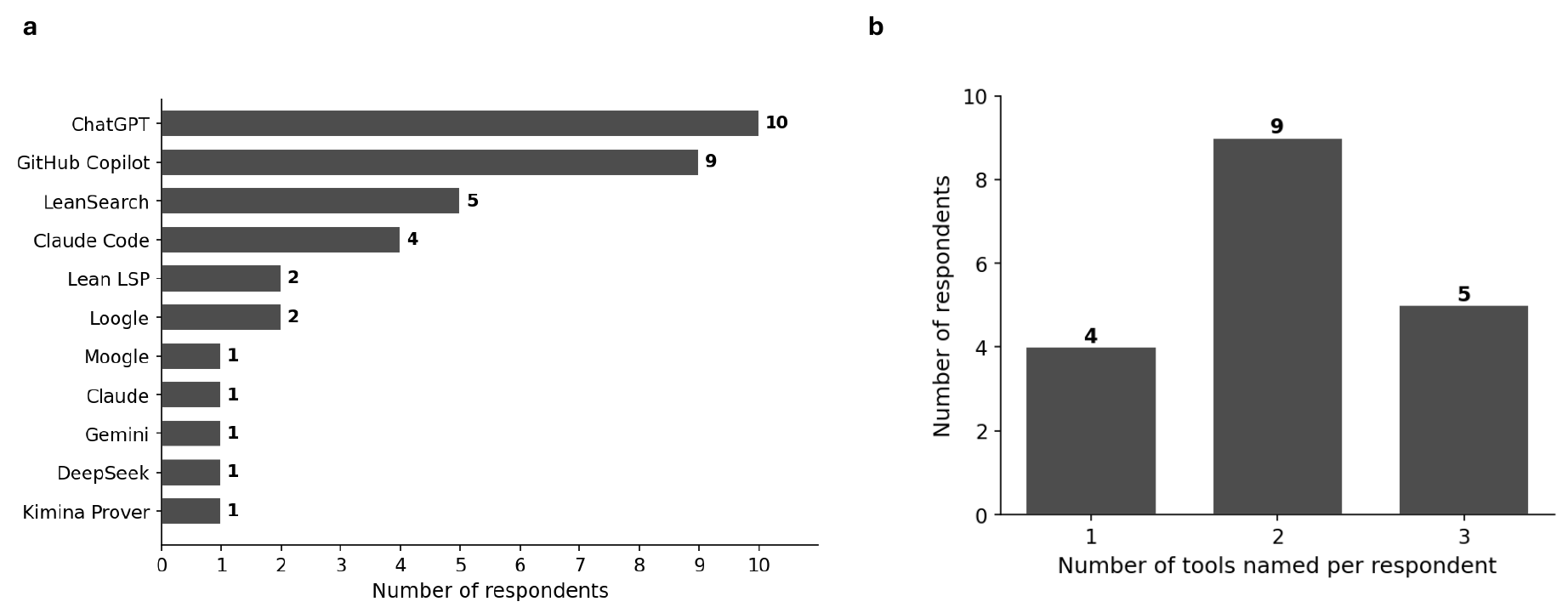}
    \caption{\textbf{Self-reported AI tool use from qualitative survey.} Respondents were prompted with example tools (taken from what participants in the user study used) and did not necessarily describe all tools they use in practice; $18$ specified at least one specific tool. \textbf{a,} Counts of tools expressed in respondents' descriptions when asked whether they used any of the same tools as participants in the user study (and to write in other tools they may use, if any). \textbf{b,} Number of distinct tools mentioned by each of the $18$ respondents who specifically named at least one explicit AI-based tool that they use.}
    \label{fig:survey-ai-tool-use}
\end{figure}

We next include additional insights into the user study. Table~\ref{tab:envisioned-workflows} shows several example survey responses. Full responses can be found at  \href{https://collinskatie.github.io/hai_formalization.github.io/}{webpage}. While we did not elicit an exhaustive set of kinds of AI tools respondents used, as in the user study, multiple respondents also indicated several different tools (Table~\ref{tab:tool-use-survey-examples}). One author (working with Claude Code) annotated the tools participants expressed\footnote{The codings are fairly coarse, as the diversity of tools continues to grow and encapsulate multiple subvariants. One participant noted they used ``Copilot-GPT-Codex.'' We coded this as GitHub Copilot}. The most common tools again were GitHub Copilot and ChatGPT (Figure~\ref{fig:survey-ai-tool-use}); however, we primed respondents with example tools (see S.I.~\ref{ref:survey-questionnaire} Q14). Even though it was not listed in our seeded example tools in the question, four respondents indicated they used Claude Code. 

\begin{figure*}
    \centering
    \includegraphics[width=1.0\linewidth]{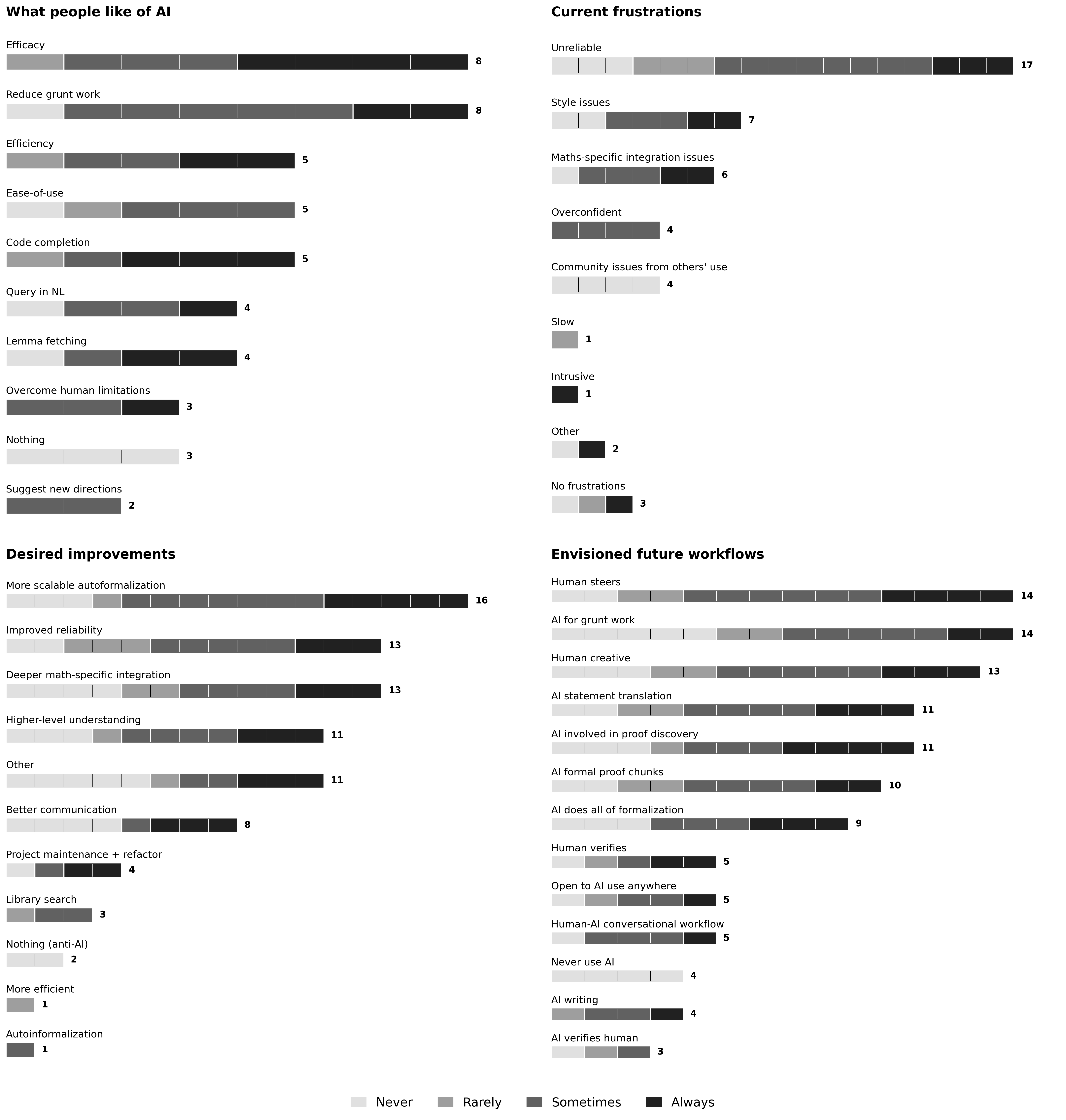}
    \caption{Survey results, decomposed by self-reported AI usage.}
    \label{fig:survey-ai-use}
\end{figure*}

\begin{figure*}
    \centering
    \includegraphics[width=1.0\linewidth]{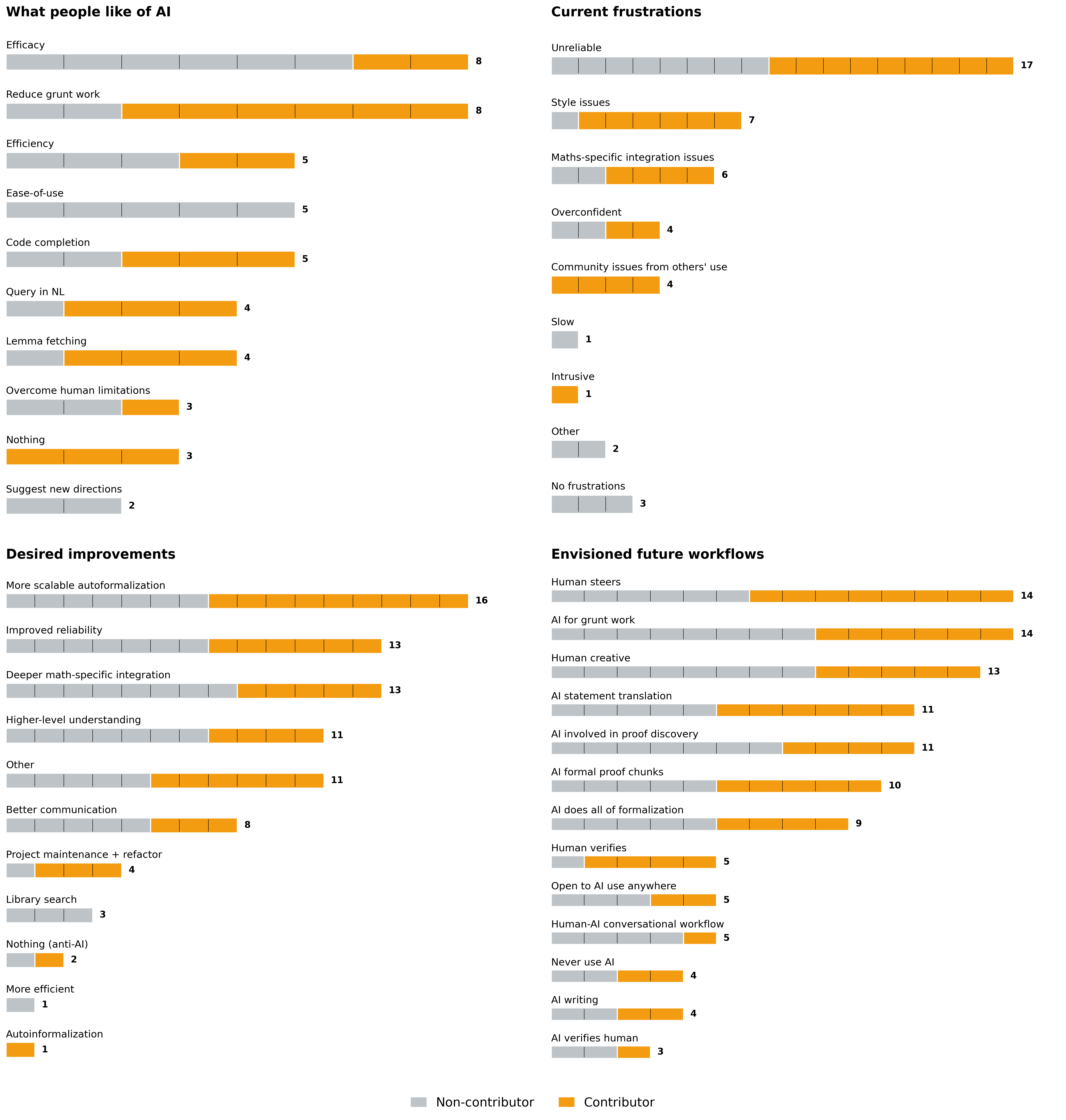}
   \caption{Survey results, decomposed by whether respondent indicated they had previously contributed to mathlib.}
    \label{fig:survey-mathlib}
\end{figure*}

\begin{figure*}
    \centering
    \includegraphics[width=1.0\linewidth]{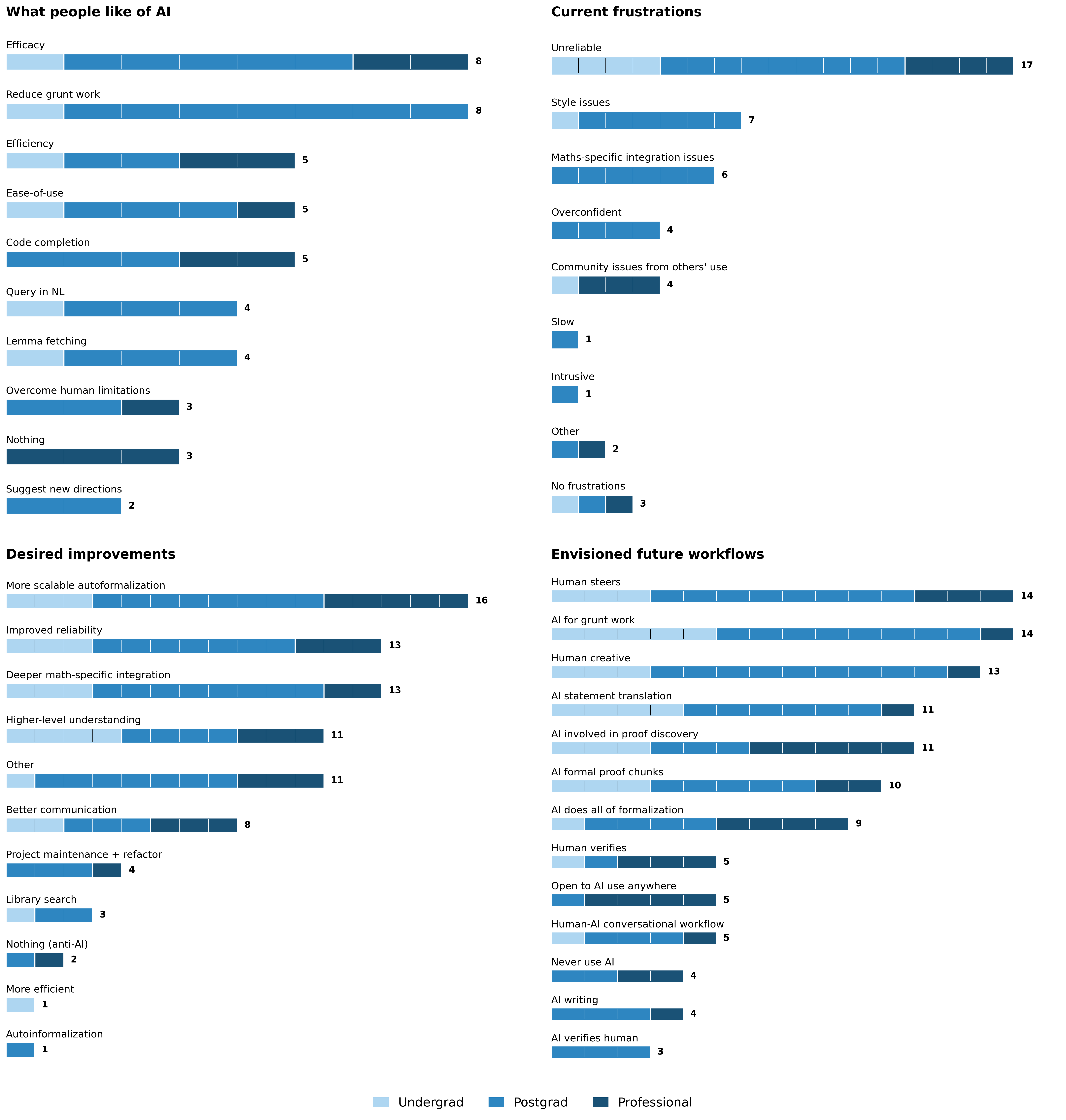}
    \caption{Survey results, decomposed by self-reported respondent general mathematics experience.}
    \label{fig:survey-math}
\end{figure*}

\begin{figure*}
    \centering
    \includegraphics[width=1.0\linewidth]{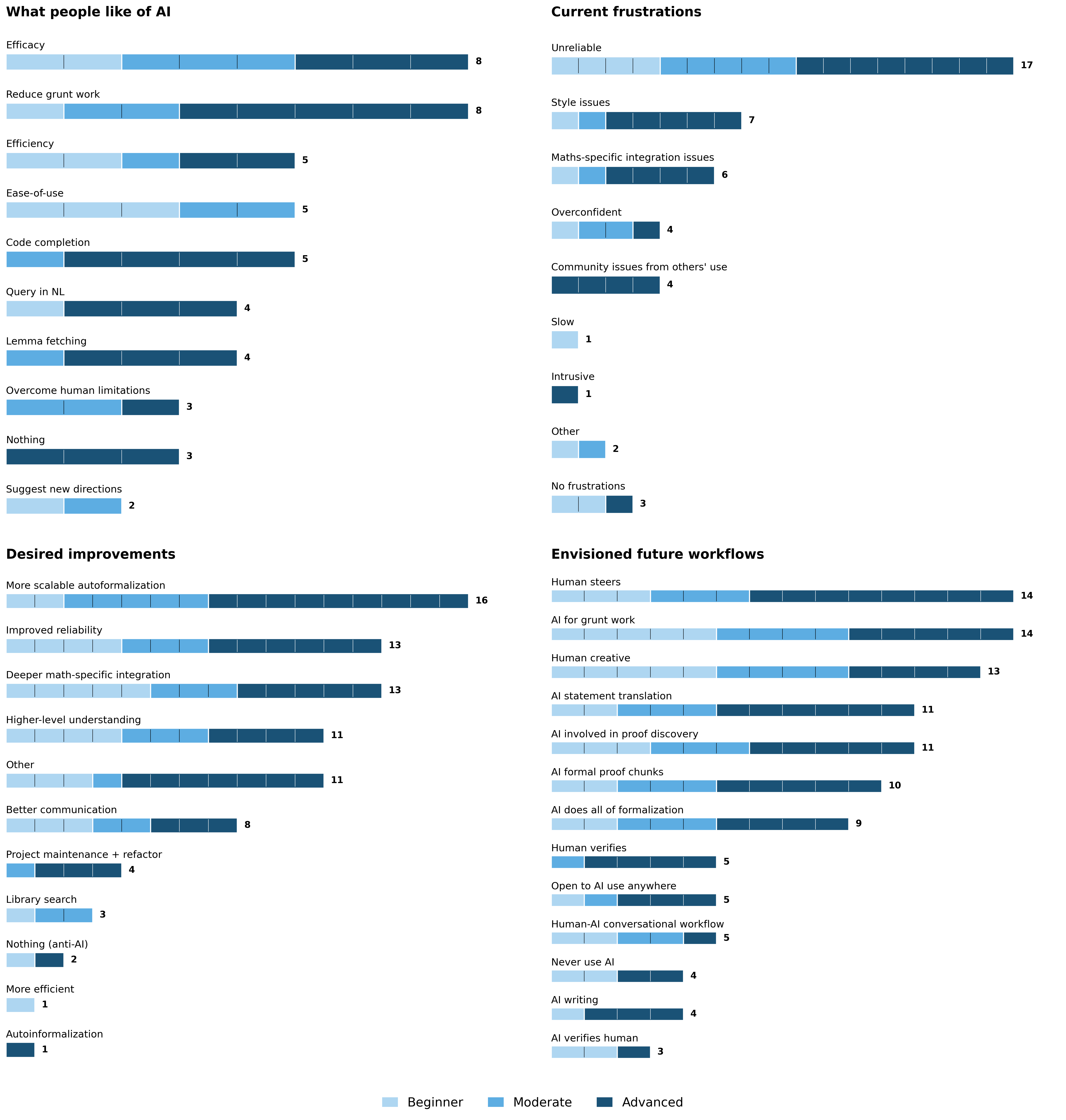}
   \caption{Survey results, decomposed by self-reported respondent formalization experience.}
    \label{fig:survey-formalization}
\end{figure*}

Additionally, we include a full breakdown of codings of responses based on respondents' self-reported experience and demographic data, i.e., their AI usage (Figure~\ref{fig:survey-ai-use}), whether they have contributed to mathlib (Figure~\ref{fig:survey-mathlib}), their mathematics experience (Figure~\ref{fig:survey-math}), and specifically their formalization experience (Figure~\ref{fig:survey-formalization}). For each, we depict codings for what respondents' expressed liking, being frustrated by, most wanting next for AI tools and what workflows they envisioned. 

\section{Additional Details on Study Design}
\label{sec:study-details}

We next provide additional details on study design, namely problem construction, participant recruitment, and problem grading.

\subsection{Additional Details on Problem Construction}
\label{sec:problem-design}

Problems were selected and constructed by our author team. All problems have sufficiently short formal proofs that a team of pilots was able to formalize all six of them in approximately twelve hours. This team was made up from the authors of this study, and formalizers all had at least a first degree in mathematics, or were at most at the level of a Post-Doc; all had significant prior experience in formalizing. 

Formalization difficulty comes in two parts: the intrinsic difficulty of formalizing a problem; and whether Mathlib provides pre-existing  definitions and useful lemmas to carry out formalization; or, at its extreme, contains an approximate version of the sought proof. In the latter case, simply adapting the key parts of the known formal proof to transform it to the sought proof is much easier than formalizing from scratch. We also account for this possibility and include both problems where a similar formalization exists, as well as problems where a similar formalization does not exist. 

We provide an overview over the problems used in our study, including their statements as they were given to participants.
For information regarding how participants performed on each problem, we refer to Section~\ref{sec:acc-granular}.

\begin{itemize}
    \item \textit{Problem N:} This is a simple number theory problem about solving a congruence equation. The expected format of the solution is deliberately open-ended to encourage participants to present their answers in diverse ways.
    \\ \textbf{Problem Statement:} \emph{Find all solutions of the congruence $12x \equiv 9 \pmod{15}$.}

    \item \textit{Problem A:} This is an analysis problem involving continuous and differentiable functions and the use of the mean value theorem. The problem is exactly Cauchy's mean value theorem, although this information was not given to participants. Cauchy's mean value theorem has been included in mathlib since 2019.
    \\ \textbf{Problem Statement:} \emph{Let $f,g: a,b]\rightarrow\mathbb{R}$ be continuous and differentiable on $(a,b)$. Then there exists $x\in(a,b)$ such that \[ ((f(b)-f(a))g'(x)=(g(b)-g(a))f'(x).\]}

    \item \textit{Problem C:} This is a counting problem about integer sequences defined by a self-referential rule involving divisibility. After classifying the solutions, one must apply foundational results about finite cardinalities, which can be cumbersome as it requires checking obvious non-equalities.
    \\ \textbf{Problem Statement:} \emph{We call a sequence $a_1, a_2, \ldots$ of non-negative integers \textit{delightful} if there exists a positive integer $N$ such that for all $n > N$, $a_n = 0$, and for all $i \geq 1$, $a_i$ counts the number of multiples of $i$ in $a_1, a_2, \ldots, a_N$. How many delightful sequences of non-negative integers are there?}

    \item \textit{Problem G:} This is a geometry problem, solved primarily using angle chasing. Part of the solution involves constructing an auxiliary point along an interval that creates a specified angle with another point, which seems ``obviously" valid to a human mathematician. Much of the difficulty in formalizing comes from showing that such a point exists, which requires careful use of the intermediate value theorem, noting that the angle subtended varies continuously as a point moves along a line.
    \\ \textbf{Problem Statement:} \emph{Let \(ABC\) be an isosceles triangle with \(AB = AC\) and
\(\angle BAC = 20^\circ\). Let \(G\) be on the segment \(AC\) such that
\(\angle ABG = 20^\circ\). Let \(H\) be on the segment \(AB\) such that
\(\angle ACH = 30^\circ\). Find \(\angle AGH\).}

    \item \textit{Problem T:} This is a topology problem involving Baire spaces, which requires unraveling definitions, and careful understanding of subspace topologies.
    \\ \textbf{Problem Statement:} \emph{A topological space $X$ is said to be ``Baire" if for any sequence $A_1, A_2,\ldots$ of open dense sets in $X$, the intersection $\bigcap_n A_n$ is dense. Suppose $X$ is a Baire space and $Y$ is an open subspace of $X$. Prove that $Y$ is Baire.}

    \item \textit{Problem V:} This is a simple visual counting problem involving tiling a 2x3 grid with dominoes. The mathematical proof is simple; the difficulty comes from having to formally state this problem, which is cumbersome to do in Lean. This problem also requires manipulation of finite cardinalities.
    \\ \textbf{Problem Statement:} \emph{Consider a $2 \times 3$ domino board. You have two $1 \times 2$ dominos and two $1 \times 1$ dominos. In how many ways can you cover the board? }

\end{itemize}

\subsection{Additional Participant Details}
\label{sec:participant-rec}

We recruited participants through our networks through university mailing lists and personal contacts. Participants were told the study would take approximately $12$ hours and were paid at a fixed rate of $\$240$ (for an estimated $\$20$/hr).  In case of questions regarding what consists of AI tool use, we worked with the participants to make sure our criteria were uniform across participants. Participants provided self-reports at the end of the study, indicating their experience with mathematics and formalization, as well as providing qualitative responses into their AI use (see Section ~\ref{sec:participant-responses}. Participants also provided screenshots of any scratchwork used during their formalization process. The study received prior ethics approval by our university departmental ethics review, and all participants provided informed consent.

All participants noted that they used Lean at least once a month (they were given the option to indicate less frequent use). All participants indicated that they had ``moderate comfort'' with at least $1$ formal proof language (Four participants were comfortable with one language; two with two languages; and one participant with three languages). Four participants indicated that they ``always'' used AI-based tools when formalizing in their day-to-day workflows; two indicated they ``sometimes'' did; and one indicated ``rarely'' using AI-based tools. 

\subsection{Additional Analysis Details}

Three experienced Lean users from our author team marked the accuracy of participants' formalized proofs. 

We code tool use based on a mixture of participants' self-reported tool use (see Table~\ref{tab:tool-use}) and notes from watching videos, as participants did not always report all tools they used. While we have watched several hours of participants' videos, and notice that most participants are consistent in their tool use across problem sessions, until we have completed a full coding of all $80$ hours of video, it is therefore possible that we missed some tool use.

\section{Additional Analyses into Participants' Formalization}
\label{sec:addtl-analyses}
\vspace{-10pt}

We include additional exploratory analyses into participants' formalization results.

\subsection{Formalization Decomposed by Experience}
\label{sec:experience}

\begin{figure}[h!]
    \centering
    \includegraphics[width=1.0\linewidth]{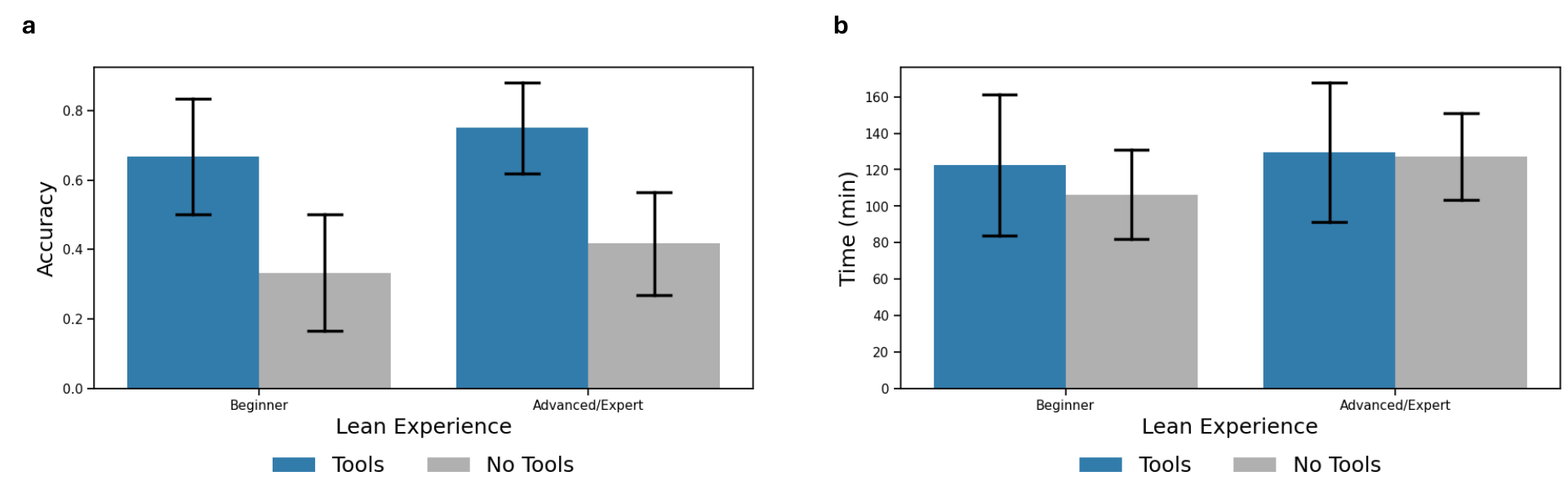}
    \caption{\textbf{Formalization performance by Lean experience.} \textbf{a,} Average accuracy based on whether participants self-reported as being Beginners ($N=3$) or Experts ($N=4$) with Lean; \textbf{b,} Average time (min) across problems. Error bars show standard error.}
    \label{fig:lean-exp}
\end{figure}

\begin{figure}[h!]
    \centering
    \includegraphics[width=1.0\linewidth]{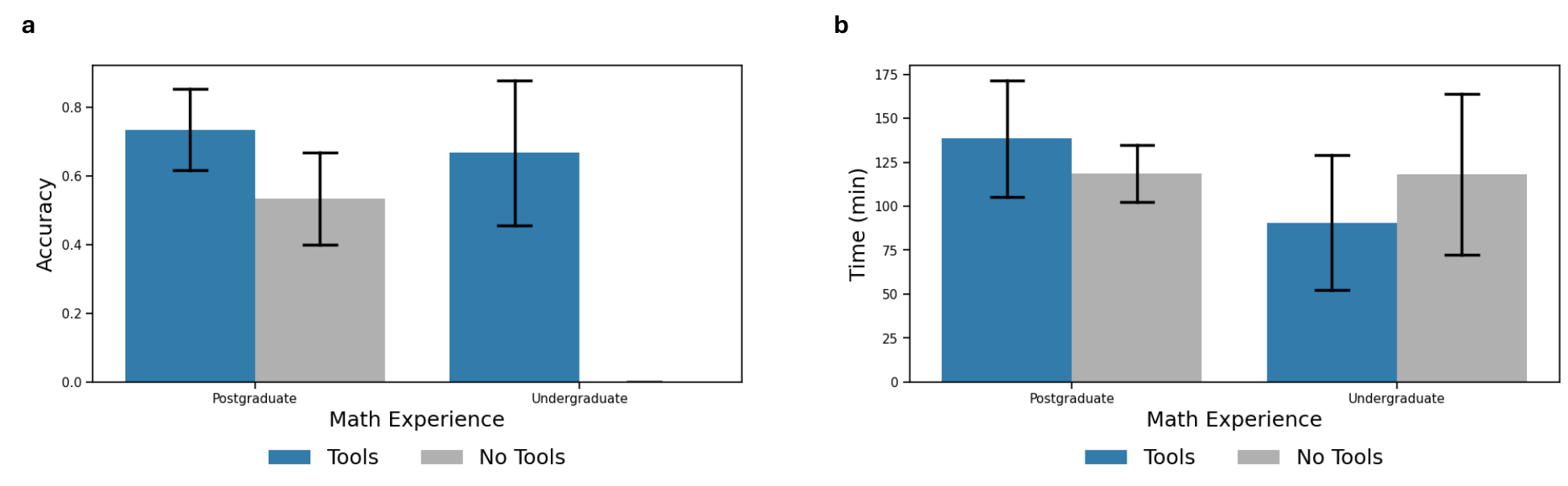}
    \caption{\textbf{Formalization performance by mathematics experience.} \textbf{a,} Average accuracy based on whether participants self-reported as having studied mathematics up to or through the postgraduate level ($N=5$) or undergraduate level ($N=2$). Accuracy is decomposed into the problems where participants did or did not have access to tools (three problems per participant per category).; \textbf{b,} Average time (min) across problems. Error bars show standard error. These analyses are descriptive and exploratory; we urge caution in any extrapolation due to the small sample size.}
    \label{fig:math-exp}
\end{figure}

We also decompose participants' formalization accuracy based on their self-reported Lean experience (Figure~\ref{fig:lean-exp}) and mathematics experience (Figure~\ref{fig:math-exp}. We caution over interpretation due to small sample sizes. 

\subsection{Finer-Grained Accuracy Evaluations}
\label{sec:acc-granular}

\begin{figure}[h!]
    \centering
    \includegraphics[width=0.7\linewidth]{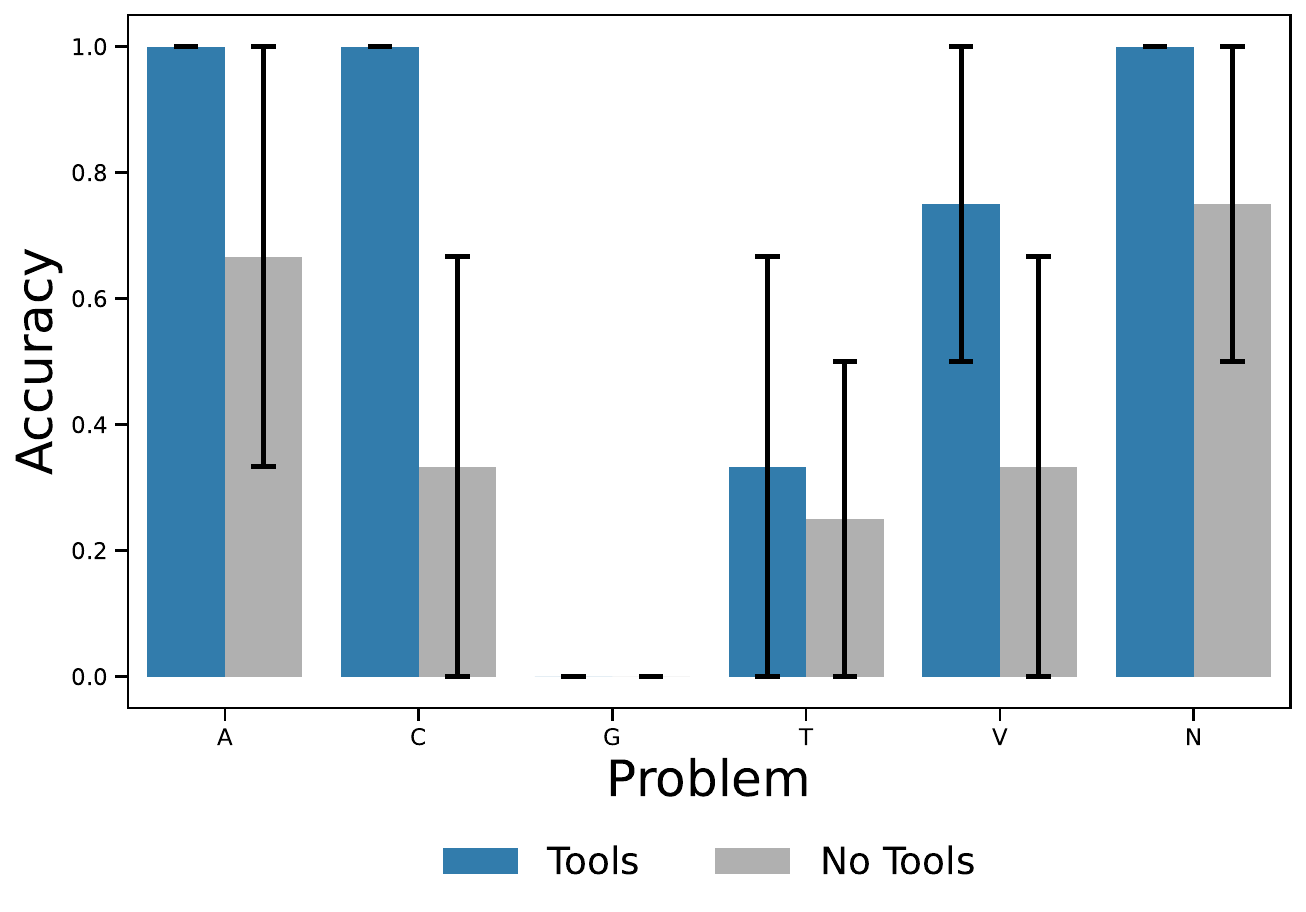}
    \caption{\textbf{Accuracy per problem, for participants with and without tool use.} Accuracy is averaged per problem. Error bars show standard error. We caution over-interpretation due to small sample sizes ($3-4$ participants) per bar.}
    \label{fig:acc-problem}
\end{figure}

Throughout the paper, we have focused on accuracy as measured by having both the statement and proof correct. We report the average accuracy per problem in Figure~\ref{fig:acc-problem}. However, a few participants were correct in their statement formalization but incorrect in the proof (or vice versa -- formalized the statement incorrectly, but proved the statement correctly under that interpretation), or formalized only part of the proof. E.g., one proof that was marked as incorrect did get some of the way there according to the grader on Problem N: 

\texttt{``Only one direction ($12x\equiv 9\pmod{15}\implies x=5n+2$) is formalized. That direction is correctly formalized and proved. \texttt{ZMOD} is used for modulus in the assumption, $5n+2$ is explicitly written as the answer.''}

Whereas another participant (for Problem G) made a start at the problem but did not formalize the entire proof correctly according to the grader: 

\texttt{``Formalised a lot, but with some sorrys (beyond existence of D). Noted that showing the existence of D ``seems hard`` ''}

We are actively expanding further exploratory analysis of a finer-grained human evaluation of the proofs. As before, three formal mathematics researchers from our team assessed each proof (one per proof). Future work can conduct a more expansive human audit of proofs. 

\subsection{Analyzing Code Structure}

\begin{figure}[h!]
    \centering
    \includegraphics[width=1.0\linewidth]{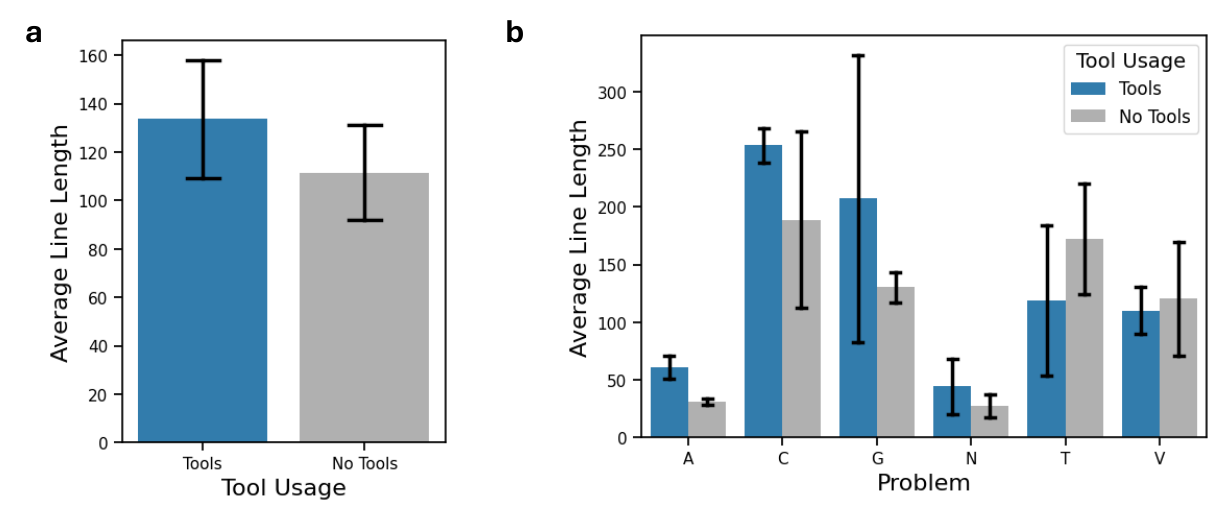}
    \caption{\textbf{Formalized proof length.} \textbf{a,} Average code length (not including comments); \textbf{b,} Average code length (not including comments) per problem.}
    \label{fig:code-len}
\end{figure}

We also conduct exploratory analyses into the style of code participants provided, depending on whether they had access to tools or not. We observe that there may be some effect of AI-based tool use on overall proof length, with participants potentially producing -- on average -- longer formal proofs with AI-assistance (Figure~\ref{fig:code-len}a). However, this effect is variable at a per-problem level (Figure~\ref{fig:code-len}b) and moderated by whether participants even finished the proof. 

\subsection{Analyses into Formalization Processes from Participant Videos}
\label{sec:video-analyses}

Participants recorded themselves formalizing proofs, producing over $80$ hours of videos. Two authors from our team manually watched and annotated a selection of videos to understand when and how participants were using AI. We are actively extending our analyses for more rigorous coding over our full library of formalization videos.  

Here, we offer a glimpse into the kinds of behavior different participants showed, with brief summaries of a few of the formalization videos. These behaviors can be characterized in varying reliance strategies in a user's AI workflow (a la~\citep{jorgensen2025documenting}). As described in the main text, one group of participants predominantly formalized the proofs themselves, only lightly engaging AI tools (often, GitHub Copilot in-line). For instance, two participants who could be classified as ``human formalizers with AI assistance'' each sparingly used GitHub Copilot as part of solving Problem V. One participant listed themselves as an expert formalizer in Lean; the other, as a Beginner Lean user. Both participants used comments to ``prompt'' GitHub Copilot and regularly overrode Copilot suggestions. One participant (with Lean experience) used Copilot in more of a coarse-to-fine fashion, working on some functions, templating them and using Copilot to fill in details, particularly with substantive and repetitive pattern matching. Relatedly, the participant with less Lean experience also regularly rejected Copilot suggestions; they can be seen frequently pausing and deliberately reading the Copilot suggestions, before occasionally accepting them -- suggestive of the intentional nature of their proof process. 
Both participants solved Problem V correctly.

In contrast, another participant could be categorized as behaving in a way of ``AI formalization, with human help.'' This participant started one problem by opening up two chat windows: one ChatGPT and one Claude. The participant then pasted in the problem PDF into each window, and proceeded to engage in a copy-output/copy-error message cycle into the models over the course of an hour. While the participant shifted to using GitHub Copilot in the latter half of the proof, even there, the participant did not seem to engage in as much critical thinking. There was a span of five minutes where the participant went back and forth with GitHub Copilot saying ``finish this,'' having the model generate a series of local code snippets (the participant queried $8$ times, accepting $7$ of them outright). This led to a substantial amount of the code not being generated by the participant; the participant seemed to recognize that the code did not match their style and noted in a comment at the top near the end of the video (``\texttt{i’m sorry this is so ugly}''). While this participant formalized the problem correctly, they took about $3\times$ as long as the average participant for this problem (``Problem N''). 

Lastly, one participant (who used nine different tools over the course of the study) was a highly varied user, engaging many different kinds of AI tools for different parts of their formalization process -- yet demonstrated regular and substantial agency in how they used the tools (e.g., regularly declined in use of AI-based tools over the course of the problem). For instance, on one problem (``Problem N''), this participant started by pasting the informal statement into Kimina to formalize, pasted into VSCode and manually edited. They then went to look at the solution, opened GitHub Copilot chat to get a tailored response using a particular tool, querying ``\texttt{Define the set of all numbers of the form x = 5n + 2 for some integer n, in Lean, using ZMOD.}'' They then further clarified their interest in ZMOD: ``\texttt{I’d like to use the ZMOD notation. An example is: [...]}'' and copied over the solution. They then asked for ``\texttt{suggest tactics}'' from Lean Copilot, looked through and chose the second suggestion, and proceeded to again search for tactics (this time choosing the third suggestion). They then accepted a major block of GitHub-generated code, before moving to Claude Sonnet to rewrite part of their code (``\texttt{I’d like to prove this just using the definition of ZMOD. Could you rewrite this proof?}''). They then proceeded to delete a large block of AI-generated code in their proof, went back to Kimina and had the model generate the full formal proof, given the informal, and made manual tweaks. However, $30$ seconds later, the participant proceeded with a series of major deletions, essentially restarting the entire formalization process and wrote the rest of the proof primarily unassisted (ending up with a much simpler proof that used the \texttt{omega} tactic). This is a case where the participant heavily engaged AI, but ended up, in their final proof, relying very little on the actual AI-generated code.

\section{Additional Participant Self-Reported Thoughts on AI-Based Assistance for Lean}
\label{sec:participant-responses} 

Participants were asked the following questions after their study (in addition to the question of what tools they used, as presented in Table~\ref{tab:tool-use}). We include participants' responses verbatim.

\subsection{``What do you like most about AI-based tools for formalizing?''}
\begin{itemize}
    \item \textit{``I like that AI tools can help me formalize "obvious" results that I have trouble formalizing on my own.  I believe that AI can also help familiarize me with new topics and good ways to formalize those topics.''}
    \item \textit{`` The can (often) take care of tedious low-level details. They also offer better "library search" than Mathlib's search function.''}
    \item \textit{``When I'm certain of an approach, I like how AI-based tools fill in the unimportant/rote details. For example, if I need to open a file in Python, rather than looking up the syntax, I can just leave a comment that I want to open a file, and Copilot will fill in the code for me. When it comes to Lean formalization, Copilot copies parts of proofs from elsewhere in the file, fetch names of lemmas from Mathlib, and fill in assumptions for lemmas, which saves me time. I rely less on its proof completion, especially for nontrivial proofs. But it works pretty well on straightforward induction proofs or simp-filled proofs.''}
    \item \textit{`` Discovering definitions and lemmas in the library is significantly easier thanks to AI-based search and autoformalization tools.''}
    \item \textit{`` I also find it helpful to load relevant Mathlib files into the GitHub Copilot chat context to ask questions that would otherwise take me a significant amount of time to find answers to.''}
    \item \textit{`` I like that it helps to get me started to have a framework of understanding the problem and giving me ideas to approach the formalization, even if it is unable to give me the full thing. It is also helpful in pointing me to other resources that I can reference''}
    \item \textit{`` leansearch gives me fast lookup with low cognitive overhead. Chatgpt/Gemini sometimes are very good at pointing out mistakes that I've made in formulating the problem.''}
    \item \textit{`` Using Copilot can save time.''}
\end{itemize}

\subsection{``What frustrates you most about the current state of AI-based tools for formalizing?''}
\begin{itemize}
   \item \textit{`` Sometimes the code is wrong or doesn't do what I want it to do (this is sometimes fixed by giving it more prompts)''}
    \item \textit{``Miscalibrated confidence in the answers. Even if the AI "doesn't know" it will output something plausible but wrong, which is more frustrating than not getting an answer at all. Also AI-based tools seem to be much worse for combinatorics and geometry than for algebra or analysis.''}
    \item \textit{``When it's wrong, it's really wrong. Sometimes Copilot suggests a (broken) "proof" of a theorem that will in reality turn out to be much longer or more complex. This requires me to cancel the suggestion and fill in the proof myself (which is what I would have done anyway, but the extra keystrokes to dismiss it can be distracting).''}
    \item \textit{``I find that there aren't enough tools specifically tailored to the Lean environment. There's been a lot of exciting progress in the AI for mathematics space in the recent years, but not all of that progress has translated into better tooling for the Lean community. Even when tools exist, they can be difficult to get working, requiring a complicated set-up involving external dependencies or an API key configuration (although it's completely understandable why this is the case). An ideal situation would be to have a VS Code extension that can interact with the current Lean session and assist the user with various formalization tasks. Another option would be to have AI-based tools and tactics shipped by default with Mathlib or Lean, but this has the disadvantage of being "opt-out" rather than "opt-in".''}
    \item \textit{``There's still quite a bit of hallucination, calling tactics that don't exist for example. Even giving it reference to the Mathlib documentation (by copying and pasting) didn't really help''}
    \item \textit{``1. Mathlib seems to evolve faster than models can keep up. 2. Chat models are often very confident about wrong things -- even stuff like "here's a much shorter simpler proof!" and it's actually longer and has error. 3. Models are expensive to use. 4. Stuff like AlphaProof is not available to the public.''}
    \item \textit{``The practice of using such tools does not match strong results reported in research benchmarks; they are mostly unable to prove even very simple things, including when given access to the Lean goal state via MCP. Also, most of the tools are challenging to set up for a local development environment.''}
\end{itemize}

\subsection{``What do you envision any future workflow pattern for how you may go about discovery new proofs / formalizing them? Where would you want tools that help you? Where would you NOT want AI-based tools to be used?''}

\begin{itemize}
    \item \textit{``Better code generation, maybe also suggest more advanced techniques (like using tactics)''}
    \item \textit{``Reduction of "bullshitting" (giving plausible but wrong answers) when they "don't know". Also, expansion of application domains to be more general than algebra and analysis.''}
    \item \textit{``Automatically fill in lemmas/definitions as I work on a larger theorem! For example, it would be great if I could leave a lemma with "sorry," and then while I use the lemma in the main theorem, Copilot (or whatever tool) could copy my file up to the lemma and work with it in the background until it finds a proof. Right now, all interactions are human-driven. (Of course, this will introduce more problems - what if the definition/proof is too verbose, or not quite what I want. Then I need to manually clean up, which is still faster than writing the proof myself, but can be more frustrating/tiring. But if it's tuned right, this would save time!)''}
    \item \textit{``Good autoformalization and proof completion models that integrate seamlessly into the Lean editor would be useful to have.''}
    \item \textit{``A generally Lean-aware chatbot that can alert the user to Lean formalization conventions (like writing `a < b' instead of `b > a') explain how a mathematical concept is defined in Mathlib, and how it is meant to be used generate outlines for proof formalization suggest a convenient way to define a mathematical concept in Lean (for example, by combining existing definitions in the library or by modelling it based on an existing definition of a similar concept) suggest new lemmas to formalize suggest attributes to tag lemmas with (like \texttt{[simp]}, \texttt{[aesop]} or \texttt{[grind]}) mention relevant lemmas, tactics or domain-specific proof formalization strategies improve and optimize proofs give high-level feedback on a Lean document (like a Mathlib reviewer) would make formalization significantly easier and more accessible, in my opinion.''}
    \item \textit{``Being able to generate tactics that actually exist. Of course ideally if it can do end to end formalization of proofs that would be amazing, but in the meantime being able to more reliably tackle smaller subproblems that contribute towards the larger goal''}
    \item \textit{``I want tools that can simplify my existing proofs.''}
    \item \textit{``Autoformalization from natural language, and automatic theorem proving.''}
\end{itemize}
    
\subsection{``What would you want to see most in future AI-based tools for formalizing proofs?''}

\begin{itemize}
    \item \textit{``If I already have an idea of the solution, I would consult a large language model for ideas on how to go about formalizing parts of the problem.  The AI tool could suggest tactics and theorems to use, and maybe give a working formalization.  I didn't try asking it for suggestions on how to solve a problem (because we already had solutions), but I think that would be cool.''}
    \item \textit{``I mostly want AI to take care of tedious low-level details of a proof; I'm happy to do the main structure.''}
    \item \textit{``It would be great if I could quickly outline the shape of the proof, such as ":= by -- induction on n, using core lemmas [X, Y, Z]" without explaining my thought process too deeply, and having the skeleton filled in. The closer a tool can allow me to type in only the key pieces, and having the rest of the proof/syntax get filled in, the better.''}
    \item \textit{``I don't like AI-based tools writing too much code for me - it makes me lose my mental model of how the program functions. But if an AI-based tool can fill in an entire proof for me, and the proof is succinct and compiles quickly, then my mental model stays intact. The problem is that AI-based proofs are often not succinct or efficient, and so I have to trade-off writing the proof correctly from the start, or editing a bad proof into a good one.''}
    \item \textit{``I think a nice workflow for formalization would be one where you could describe a mathematical idea or proof gradually to a chatbot and get suggestions for ways to formalize it in Lean. In this interaction, the human only needs to have a passing familiarity with the Lean syntax, and a bulk of the work will be done by the machine. To keep the response time reasonable, ideally the chatbot should not spend an enormous amount of time autonomously formalizing or discovering proofs, but instead should develop these in conversation with the human.''}
    \item \textit{``I think I would use it quite regularly for generating small tactics that help me do the mundane stuff, but overall I probably would still have to guide the overall structure of the proof and think of next steps''}
    \item \textit{``More real-time feedback, in an unobstrusive way. Interactive theorem proving strongly relies on using goal states and other feedback from the theorem prover. AI-based workflows should integrate better with this information and use the information rather than trying to one-shot the entire proof.''}
\end{itemize}

\section{Survey Questionnaire}
\label{ref:survey-questionnaire}

We provide the full survey questionnaire circulated to respondents. The survey unfolded over three pages: a brief introduction, the consent form, and the main questionnaire. 

\subsection*{Introduction:}
\begin{mdframed}[backgroundcolor=gray!10, linecolor=gray!50, linewidth=0.5pt, roundcorner=3pt, skipabove=0pt, skipbelow=0pt, innertopmargin=8pt, innerbottommargin=8pt]
\textbf{AI Tool Use in Formalization Processes}
 
\medskip
 
We are interested in understanding how people do (or don't) use AI-based tools as part of their formalization workflows. Thank you in advance for filling out this survey!
 
Three randomly-selected participants will receive a \$30 Amazon gift card!
 
The survey should take no more than 5--10 min.
 
Responses from the survey will be analyzed by us and other researchers to better understand how people are using AI-based tools (or not using them!) as part of formalization. Emails are collected only to reduce spammers and to contact gift card winners. Emails will be discarded after the survey collection and not shared in any future data analysis. Only open-ended commentary responses and experience questions (de-linked from emails) may be shared with our later publication.
 
There will first be a brief consent form on data sharing and then the survey!
 
Any questions? \texttt{kmc61@cam.ac.uk} and \texttt{simon.frieder@cs.ox.ac.uk}
 
\medskip
 
$^*$ Indicates required question
 
\medskip
 
\textbf{Q1.} Email $^*$
\end{mdframed}
 
\subsection*{Consent form:}
\begin{mdframed}[backgroundcolor=gray!10, linecolor=gray!50, linewidth=0.5pt, roundcorner=3pt, skipabove=0pt, skipbelow=0pt, innertopmargin=8pt, innerbottommargin=8pt]
This survey is for research purposes.
 
By completing this survey, you are participating in a study being performed by researchers from the University of Cambridge and other universities (e.g., CMU, Caltech, NYU). The purpose of this research is to evaluate the role of AI and non-AI tool-based assistance in formalization.
 
You must be at least 18 years old to participate. There are neither specific benefits nor anticipated risks associated with participation in this study. Your participation in this study is completely voluntary and you can withdraw at any time by simply exiting the study. You may decline to answer any or all of the following questions. Choosing not to participate or withdrawing will result in no penalty. Your anonymity is assured; the researchers who have requested your participation will not receive any personal information about you, and any information you provide will not be shared in association with any personally identifying information.
 
We will analyze and release the free form text responses, as well as the other question data (emails will not be included!) Please do not participate unless you are okay with the responses potentially being shared.
 
You may print a copy of this consent form for your records.
 
If you have questions about this research, please contact the researchers by sending an email to \texttt{kmc61@cam.ac.uk}. These researchers will do their best to communicate with you in a timely, professional, and courteous manner.
 
If you have questions regarding your rights as a participant, or if problems arise which you do not feel you can discuss with the researchers, please contact the University of Cambridge Dept of Engineering Ethics Offices.
 
\medskip
 
\textbf{Q2.} I am age 18 or older $^*$ \quad (Yes; No)
 
\medskip
 
\textbf{Q3.} I have read and understand the information above. $^*$ \quad (Yes; No)
 
\medskip
 
\textbf{Q4.} I want to participate in this research and continue with the survey. $^*$ \quad (Yes; No)
\end{mdframed}
 
\subsection*{Main questionnaire:}
\begin{mdframed}[backgroundcolor=gray!10, linecolor=gray!50, linewidth=0.5pt, roundcorner=3pt, skipabove=0pt, skipbelow=0pt, innertopmargin=8pt, innerbottommargin=8pt]

Onto the survey! 

Now here are a few questions (should not take more than 5--10 min). Thank you for your time!
 
\medskip
 
\textbf{Q5.} How would you classify the amount of formalization experience you have (in any formalization language, e.g., Lean, Rocq, Isabelle)? $^*$
 
(Early beginner; Beginner; Moderate; Advanced; Expert)
 
\medskip
 
\textbf{Q6.} How would you classify your level of (specifically) Lean experience? $^*$
 
(Early beginner; Beginner; Moderate; Advanced; Expert)
 
\medskip
 
\textbf{Q7.} How often do you use *any* formal proof language (Lean, Isabelle, Rocq, and/or others)? $^*$
 
(I've never used a proof formalization language; I've only used a proof formalization language once; Once a month; A few times a month; Once a week; A few times a week; Every day; Multiple times a day)
 
\medskip
 
\textbf{Q8.} Provide a short textual description for the projects in the previous two questions. If they coincide, please state so. $^*$
 
(Open-ended text response)
 
\medskip
 
\textbf{Q9.} How often do you use (specifically) Lean? $^*$
 
(I've never used Lean; I've only used Lean once; Once a month; A few times a month; Once a week; A few times a week; Every day; Multiple times a day)
 
\medskip
 
\textbf{Q10.} How many proof formalization languages would you say you have moderate comfort with? $^*$
 
(0 (None); 1; 2; 3; 4; 5; 6; 7; $>$ 7)
 
\medskip
 
\textbf{Q11.} What is the highest level of mathematics you studied attained? $^*$
 
(High school level maths; Undergraduate level maths (non-major, but took courses); Undergraduate level maths (major); Postgraduate (Masters); Postgraduate (PhD); Professional/working mathematician; Other)
 
\medskip
 
\textbf{Q12.} Have you ever contributed to the Lean mathematics library Mathlib? $^*$ \quad (Yes; No)
 
\medskip
 
\textbf{Q13.} How often do you find yourself using AI-based tools when working on formalization? $^*$
 
(I always use AI-based tools when formalizing; I sometimes use AI-based tools when formalizing; I rarely use AI-based tools when formalizing; I never use AI-based tools when formalizing)
 
\medskip
 
\textbf{Q14.} If you use AI-based tools (e.g., GitHub Copilot, ChatGPT, Claude, Lean Copilot, LeanSearch, LeanAide, Moogle, Kimina Prover) in your formalization process, which tools do you use and in what way do you use each tool? Please write at least one bullet for each tool and method of use.
 
If you never use AI-based tools when formalizing, please write why not (e.g. unaware of them). $^*$
 
(Open-ended text response)
 
\medskip
 
\textbf{Q15.} What do you like most about AI-based tools for formalizing (answer with respect to one, or a few tools, that you mentioned above) $^*$
 
(Open-ended text response)
 
\medskip
 
\textbf{Q16.} What frustrates you most about the current state of AI-based tools for formalizing? $^*$
 
(Open-ended text response)
 
\medskip
 
\textbf{Q17.} What would you want to see most in future AI-based tools for formalizing proofs? $^*$
 
(Open-ended text response)
 
\medskip
 
\textbf{Q18.} What do you envision any future workflow pattern for how you may go about discovery new proofs / formalizing them? $^*$
 
(Open-ended text response)
 
\medskip
 
\textbf{Q19.} Where would you want tools that help you? Where would you NOT want AI-based tools to be used? $^*$
 
(Open-ended text response)
 
\medskip
 
\textbf{Q20.} Anything else you'd like to add?
 
(Open-ended text response)
\end{mdframed}

\section{Participant Instructions}
\label{ref:participant-instructions}

Participants in our controlled user study were given the following instructions and told which order to do their TOOLS or NO-TOOLS weeks, as outlined above. The instructions included links to the consent form and post-survey. Participants were sent instructions via email. Following a few questions from participants, we sent a clarifying email on tools, as outlined below.

\subsection*{Instructions}
\begin{mdframed}[backgroundcolor=gray!10, linecolor=gray!50, linewidth=0.5pt, roundcorner=3pt, skipabove=0pt, skipbelow=0pt, innertopmargin=8pt, innerbottommargin=8pt]
\begin{enumerate}
    \item Please formalize the following three problems over the course of this week. This includes both the formalization of the problem statement and the solution.
    \item When formalizing the problem statement, aim to stay as close as possible to the original statement.
    \item For the solution, you may either follow the provided approach or develop your own, as long as it leads to a correct and complete formal proof. We have formalized the solutions ourselves, so you can rest assured that the problems are formalizable without undue effort.
    \item Don't worry if you cannot finish any problem. You can work on the problems in any order, including starting one problem, stopping, and starting another before coming back to a problem. We care about your process! We what do ask you, is to make sure that you submit \emph{a single screen recording per problem} and not mix work on multiple problems within a single screen recording.
    \item If this is assigned for your \textbf{NO-TOOLS week}, please DO NOT use any external tools for assistance in formalizing the proof. If you use an IDE, such as Visual Studio Code, make sure to also turn off any GitHub Copilot or related tools within that environment.
    \item If this is your \textbf{TOOLS-ALLOWED week}, you are allowed to use any AI-based tool(s) for assistance (this includes LLMs, tools specific to the Lean ecosystem, as well as any other tools you might find useful).
    \item We kindly ask that whenever you are formalizing the problems, at any time, please record your ENTIRE screen (e.g., using QuickTime video, vokoscreenNG or any other tool you might find useful). We ask that you record \emph{all} interactions related to your problem solving; this include web browsing (if you use Google search, or LLMs), as well as any other software that assists you in formalizing. If you ever forget to record your screen, please note this in the material that you submit back as part of the study. While it is possible that you may forget, if you do this more than twice, you may not be paid. Do not worry if you accidentally screen-record personal matters - there is an option to black these out at the end, to exclude any frames that reveal personal information.
\end{enumerate}
\end{mdframed}

\subsubsection*{Submission Details}
\begin{mdframed}[backgroundcolor=gray!10, linecolor=gray!50, linewidth=0.5pt, roundcorner=3pt, skipabove=0pt, skipbelow=0pt, innertopmargin=8pt, innerbottommargin=8pt]
At the end of the study, please \textbf{submit a zip} with the following information: 

\begin{itemize}
    \item Your formalized proofs for each problem – in a file labeled ``\{your-last-name\}\-\{your-first-name\}\-\{problem-id\}.txt''
    \item Any screen recording for that problem labeled ``\{your-last-name\}\-\{your-first-name\}\-\{problem-id\}-\{record-idx\}.mp4'' for each record-idx (between 0 to however many clips you took for that problem)
    \item A  \href{https://forms.gle/KBX6Ec3TS5LpfBGfA}{Google form} with your post-survey responses, including a general questionnaire about your prior mathematics and Lean experience.
    \item If you use scratch paper at any time during the study, we ask that you screenshot or upload the scratchpaper with the label ```\{your-last-name\}\-\{your-first-name\}\-\{problem-id\}-\{page-idx\}.png/pdf''
\end{itemize}

Any / all of the information may be released as part of the data collected in this work. However, we will scrub any names so the data is anonymized. We will then stitch these videos together. We will make the screen recordings public (and anonymized) as part of the study contributions; however, you will be given a chance to ``mask'' any part of the recording before publication of your accidentally screen-recorded material that should not be made public.

By participating, you agree to having data shared and affirming that you are at least 18 years old. We will communicate payment details after the study. \textbf{Please make sure you have filled out the \href{https://forms.gle/MQ3VpZjNLatkmRc68}{consent form} (also emailed) before you begin.} You only need to fill out the consent form once.
\end{mdframed}

\subsection*{Clarification email excerpts}
\begin{mdframed}[backgroundcolor=gray!10, linecolor=gray!50, linewidth=0.5pt, roundcorner=3pt, skipabove=0pt, skipbelow=0pt, innertopmargin=8pt, innerbottommargin=8pt]
We wanted to provide some clarification on what counts as a ``tool`` for the study, following a great question by a participant.
In our instructions we said: ``If this is assigned for your NO-TOOLS week, please DO NOT use any external tools for assistance in formalizing the proof.'' but we realized it might not be clear to everyone where exactly to draw the line what a tool is and what not - we received some questions from some of you about this, so we wanted to write an email to everyone to make sure we are all aligned:
If you use Google search, or any other ``tool'' that is neither specifically designed to aid math and formalization, this counts as NO TOOL, the only exception being LLMs, which do count as TOOL (not all LLMs were are designed to aid math and formalization, but still provide significant support, which is why we categorized them as tools).
We don't want write a list of potential tools here, since we do not want to bias you. But we want to encourage you nonetheless to ask us beforehand, so if you are unsure, whether something counts as tool or not - PLEASE ASK US FIRST.
\end{mdframed}

\begin{mdframed}[backgroundcolor=gray!10, linecolor=gray!50, linewidth=0.5pt, roundcorner=3pt, skipabove=0pt, skipbelow=0pt, innertopmargin=8pt, innerbottommargin=8pt]
You are not expected to spend more than 12 hrs on the problems over the two weeks. We recognize some problems are more difficult than others, and may take longer. You may not be able to finish all problems in 12 hrs! We encourage you to make sure you at least attempt all problems (rather than spend all your time on a subset and miss out on trying one or more problems). Of course if you want to spend more than 12 hours, awesome!, but that is not expected at all. 
\end{mdframed}
%TC:ignore
\end{document}